\pdfoutput=1

\documentclass[preprint]{elsarticle}

\usepackage[english]{babel}

\usepackage{hyperref}
\usepackage{url}
\usepackage{multirow}
\usepackage{booktabs}       
\usepackage{amsfonts}       
\usepackage{nicefrac}       
\usepackage{microtype}      
\usepackage{hyperref}
\usepackage{multirow}
\usepackage{multicol}
\usepackage{color}
\usepackage{enumitem}
\usepackage{graphicx} 
\usepackage{subfigure}
\usepackage{wrapfig}
\usepackage{amsmath, amssymb}
\usepackage[font=small,labelfont=bf]{caption}
\usepackage{setspace}
\usepackage{calc}
\usepackage{tikz}
\usepackage{amsmath, amssymb,bm,color}
\usepackage{marginnote}
\usepackage{adjustbox}
\usepackage{tabularx} 
\usepackage{ragged2e}
\usepackage{longtable}
\usepackage{siunitx,etoolbox}
\usepackage{afterpage}
\usepackage{rotating}

\usetikzlibrary{positioning,arrows,calc}

\newcommand{\field}[1]{\mathbb{#1}}
\newcommand{\R}{\field{R}} 

\newcommand{\vct}[1]{\boldsymbol{#1}} 
\newcommand{\mat}[1]{\boldsymbol{#1}} 

\newcommand{\tabincell}[2]{\begin{tabular}{@{}#1@{}}#2\end{tabular}}
\newcommand{\rowcell}[2]{\multirow{#1}{*}{\parbox{\hsize}{#2}}}

\robustify\bfseries

\journal{Journal of Biomedical Informatics}

\begin{document}
	
\begin{frontmatter}
	
    \title{Benchmark of Deep Learning Models on Large Healthcare MIMIC Datasets}
	
	\author[usc]{Sanjay Purushotham\corref{coauthor}}
	\cortext[coauthor]{Co-first authors.}
	\ead{spurusho@usc.edu}
	
	\author[thu]{Chuizheng Meng\corref{coauthor}}
	\ead{mengcz95thu@gmail.com}
	
	\author[usc]{Zhengping Che}
	\ead{zche@usc.edu}
	
	\author[usc]{Yan Liu}
	\ead{yanliu.cs@usc.edu}
	
	\address[usc]{University of Southern California, Los Angeles, CA 90089, US}
	\address[thu]{Tsinghua University, Beijing 100084, China}

    \begin{abstract}
Deep learning models (aka Deep Neural Networks) have revolutionized many fields including computer vision, natural language processing, speech recognition, and is being increasingly used in clinical healthcare applications. However, few works exist which have benchmarked the performance of the deep learning models with respect to the state-of-the-art machine learning models and prognostic scoring systems on publicly available healthcare datasets. In this paper, we present the benchmarking results for several clinical prediction tasks such as mortality prediction, length of stay prediction, and ICD-9 code group prediction using Deep Learning models, ensemble of machine learning models (Super Learner algorithm), SAPS II and SOFA scores. We used the Medical Information Mart for Intensive Care III (MIMIC-III) (v1.4) publicly available dataset, which includes all patients admitted to an ICU at the Beth Israel Deaconess Medical Center from 2001 to 2012, for the benchmarking tasks. 
Our results show that deep learning models consistently outperform all the other approaches especially when the `raw' clinical time series data is used as input features to the models.
\end{abstract}
    	
	
\end{frontmatter}

	
	\section{Introduction}
	\label{sec:introduction}
	Quantifying patient health and predicting future outcomes is an important problem in critical care research. Patient mortality and length of hospital stay are the most important clinical outcomes for an ICU admission, and accurately predicting them can help with the assessment of severity of illness; and determining the value of novel treatments, interventions and health care policies.
With the goal of accurately predicting these clinical outcomes, researchers have developed novel machine learning models~\citep{rel4,harutyunyan2017multitask} and scoring systems~\citep{le1993new} while measuring the improvement using performance measures such as sensitivity, specificity and Area under the ROC (AUROC). The availability of large healthcare databases such as Medical Information Mart for Intensive Care (MIMIC-II and III) databases~\citep{lee2011open,johnson2016mimic} has accelerated the research in this important area as evidenced by a lot of recent publications~\citep{johnsonreproducibility,che2015deep,che2016interpretable,harutyunyan2017multitask,rel1,rel2,rel3,rel4,rel5,rel6,rel7,rel8,rel9,rel10,rel11,rel12,rel13,rel14,rel15}.

Severity scores such as SAPS-II~\citep{le1993new}, SOFA~\citep{vincent1996sofa}, and APACHE~\citep{knaus1981apache} have been developed with the objective of predicting hospital mortality from baseline patient characteristics, defined as the measurements obtained within the first 24 hours after ICU admission. Most of these scoring systems choose a small number of hand-picked explanatory predictors and use simple models such as logistic regression to predict mortality, while making linear and additive relationship assumptions between the outcome variable (mortality) and the predictors. Earlier studies~\citep{dybowski1996prediction,kim2011comparison} have shown that such assumptions are unrealistic and that nonparametric methods might perform better than standard logistic regression models in predicting ICU mortality.

With the recent advances and success of machine learning and deep learning, many researchers have adopted these models for clinical prediction tasks for ICU admissions.
Early works~\citep{tu1993use,doig1993modeling,hanson2001artificial} showed that machine learning models obtain good results on mortality prediction and forecasting length of stay in ICU.
Recently, \citet{pirracchio2016mortality} showed that a Super Learner algorithm~\citep{polley2010super}-an ensemble of machine learning models, offers improved performance for predicting hospital mortality in ICU patients and compared its performance to several severity scores on the MIMIC-II dataset.
\citet{johnsonreproducibility} compared several published works against gradient boosting and logistic regression models using a simple set of features extracted from MIMIC-III dataset~\citep{johnson2016mimic} for ICU mortality prediction. \citet{harutyunyan2017multitask} empirically validated four clinical prediction benchmarking tasks on the MIMIC-III dataset using deep models. Even though some of these recent efforts have attempted to benchmark the machine learning models on MIMIC datasets, they do not provide a consistent and exhaustive set of benchmark comparison results of deep learning models for a variety of prediction tasks on the large healthcare datasets.
Thus, in this paper, we report an exhaustive set of benchmarking results of applying deep learning models for MIMIC-III dataset and compare it with state-of-the art machine learning approaches and scoring systems. Table~\ref{tab:compare-benchmark} shows the comparison of benchmarking works. We summarize the main contributions of this work below:

\begin{itemize}
	\item We present detailed benchmarking results of deep learning models on MIMIC-III dataset for three clinical prediction tasks including mortality prediction, forecasting length of stay, and ICD-9 code group prediction. Our experiments show that deep learning models consistently perform better than the several existing machine learning models and severity scoring systems.
	\item We present benchmarking results on different feature sets including `processed' and `raw' clinical time series. We show that deep learning models obtain better results on `raw' features which indicates that rule-based preprocessing of clinical features is not necessary for deep learning models.
\end{itemize}

The remainder of this paper is arranged as follows:
in Section~\ref{sec:related}, we provide an overview of the related work; in Section~\ref{sec:mimiciii}, we describe MIMIC-III dataset and the pre-processing steps we employed to obtain the benchmark datasets; the benchmarking experiments is discussed in Section~\ref{sec:experiments}; and we conclude with summary in Section~\ref{sec:summary}.

\begin{table}[hbt]
    \renewcommand{\arraystretch}{1.2}
	\centering
	\caption{Comparison of benchmarking works.}
	\resizebox{\textwidth}{!}{%
		\begin{tabular}{@{}clcccc@{}}
			\toprule
			&    & \citeauthor{pirracchio2016mortality} & \citeauthor{harutyunyan2017multitask} & \citeauthor{johnsonreproducibility} & \multirow{2}{*}{\textbf{This Work}} \\
            &    & \citeyear{pirracchio2016mortality} & \citeyear{harutyunyan2017multitask} & \citeyear{johnsonreproducibility} & \\ \midrule
			\multirow{2}{*}{\tabincell{c}{Time \\ Durations}}
            & 24 hours & \checkmark  &    & \checkmark  & \checkmark \\
			& 48 hours &    & \checkmark  & \checkmark & \checkmark \\ \midrule
			\multirow{2}{*}{\tabincell{c}{Number of \\ Features}}
            & Smaller feature set & \checkmark  & \checkmark & \checkmark  & \checkmark \\
			& Larger feature set &    & &    & \checkmark \\ \midrule
			\multirow{2}{*}{\tabincell{c}{Feature \\ Type}}
			& Non-time series & \checkmark  &  & \checkmark  & \checkmark \\
			& Time-series &    & \checkmark &    & \checkmark \\ \midrule
			\multirow{3}{*}{Databases}
            & MIMIC-II & \checkmark  &    & & \\
			& MIMIC-III &    & \checkmark  & \checkmark & \checkmark \\
			& MIMIC-III (CareVue) &    & &    & \checkmark \\ \midrule
			\multirow{2}{*}{\tabincell{c}{Scoring \\ Systems}}
            & SAPS -II & \checkmark  &    & & \checkmark \\
			& SOFA & \checkmark  &    & & \checkmark \\ \midrule
			\multirow{2}{*}{\tabincell{c}{Prediction \\ Algorithms}}
            & Machine learning models & \checkmark  &    & \checkmark  & \checkmark \\
			& Deep learning models &    & \checkmark  &    & \checkmark \\ \midrule
			\multirow{6}{*}{\tabincell{c}{Prediction \\ Tasks}}
            & In-hospital mortality & \checkmark  & \checkmark & \checkmark  & \checkmark \\
			& Short-term mortality &    & &    & \checkmark \\
			& Long-term mortality &    & &    & \checkmark \\
			& Length of stay &    & \checkmark  &    & \checkmark \\
			& Phenotyping&    & \checkmark  &    & \\
			& ICD-9 code group&    & \checkmark  &    & \checkmark \\ \bottomrule
		\end{tabular}
	}
\label{tab:compare-benchmark}
\end{table}

	\section{Related Work}
	\label{sec:related}
	
We first provide a brief review of machine learning and deep learning models for healthcare applications, and then discuss the existing works on benchmarking healthcare datasets.

Early works~\citep{caruana1996using, cooper1997evaluation} have shown that machine learning models obtain good results on mortality prediction and medical risk evaluation. Physionet challenge\footnote{\url{https://physionet.org/challenge/}} - a friendly competition platform - has resulted in development of machine learning models for addressing some of the open healthcare problems. With the recent advances in deep learning techniques, there is a growing interest in applying these techniques to healthcare applications due to the increasing availability of large-scale health care data~\citep{lasko2013computational,che2015deep,oellrich2015digital,che2015distilling}. For example, \citet{che2015deep} developed a scalable deep learning framework which models the prior-knowledge from medical ontologies to learn clinically relevant features for disease diagnosis. A recent study~\citep{dabek2015neural} showed that a neural network model can improve the prediction of several psychological conditions such as anxiety, behavioral disorders, depression, and post-traumatic stress disorder. Other recent works~\citep{hammerla2015pd,lipton2015learning,purushotham2016variational} have leveraged the power of deep learning approaches to model diseases and clinical time series data. These previous work have demonstrated  the strong performance by deep learning models in health care applications, which significantly alleviates the tedious work on feature engineering and extraction.

The availability of deidentified public datasets such as Medical Information Mart for Intensive Care (MIMIC-II~\citep{lee2011open} and MIMIC-III~\citep{johnson2016mimic}) has enabled researchers to benchmark machine learning models for studying ICU clinical outcomes such as mortality and length of hospital stay.
\citet{pirracchio2016mortality} used MIMIC II clinical data~\citep{lee2011open} to predict mortality in the ICU and showed that the Super Learner algorithm - an ensemble of machine learning models, performs better than SAPS II, APACHE II and SOFA scores. Their work showed that machine learning models outperform the prognostic scores, but they did not compare their results with the recent deep learning models.
\citet{harutyunyan2017multitask} proposed a deep learning model called multi-task Recurrent Neural Networks to empirically validate four clinical prediction benchmarking tasks on the MIMIC-III database. While, their work showed promising benchmark results of deep learning models, they  compared their proposed model only with a standard Logistic Regression model and a Long Short Term Memory Network~\citep{hochreiter1997long}, and omitted comparison with scoring systems (SAPS-II) or other machine learning models (such as Super Learner).
\citet{johnsonreproducibility} studied the challenge of reproducing the published results on the public MIMIC-III dataset using a case-study on mortality prediction task. They reviewed 28 publications and then compared
the performance reported in these studies against gradient boosting and logistic regression
models using a simple set of features extracted from MIMIC-III dataset. They demonstrated that the large
heterogeneity in studies highlighted the need for improvements in the way that prediction tasks are reported to enable fairer comparison between models.
Our work advances the efforts of these previous benchmark works by providing a consistent and exhaustive set of benchmarking results of deep learning models on several prediction tasks.

	\section{MIMIC-III Dataset}
	\label{sec:mimiciii}
	In this section, we describe the MIMIC-III dataset and discuss the steps we employed to preprocess and extract the features for our benchmarking experiments.

\subsection{Dataset Description}
\label{sec:mimiciii-description}
MIMIC III~\citep{johnson2016mimic} is a publicly available critical care database maintained by the Massachusetts Institute of Technology (MIT)'s Laboratory for Computational Physiology. This database integrates deidentified, comprehensive clinical data of patients admitted to an Intensive Care Unit (ICU) at the Beth Israel Deaconess Medical Center (BIDMC) in Boston, Massachusetts during 2001 to 2012.

MIMIC-III contains data associated with \num{53423} distinct hospital admissions for adult patients (aged 15 years or above) and \num{7870} neonates admitted to an ICU at the BIDMC. The data covers \num{38597} distinct adult patients with \num{49785} hospital admissions. To obtain consistent benchmarking datasets, in this paper we only include the first ICU admission of the patients.
Table~\ref{tab:dataset_statistics} shows the statistics of our dataset, and Table~\ref{tab:baseline_characteristics} shows the baseline characteristics and outcome measures of our dataset. We observe that the median age of adult patients is \SI{65.86} years (Quartile Q1 to Q3: \numrange{52.72}{77.97}) with \SI{56.76}{\percent} patients are male, in-hospital mortality around \SI{10.49}{\percent} and the median length of an hospital stay is \num{7.08} days (Q1 to Q3: \numrange{4.32}{12.03}).

\begin{table}[htb]
    \renewcommand{\arraystretch}{1.2}
	\centering
	\caption{Summary statistics of MIMIC-III dataset.}
	\resizebox{\textwidth}{!}{%
    	\begin{tabular}{lS}
    		\toprule
    		\textbf{Data} & \textbf{Total} \\ \midrule
    		\# admissions in the MIMIC-III (v1.4) database & \num{58576} \\
    		\# admissions which are the first admission of the patient & \num{46283} \\
    		\# admissions which are the first admission of an adult patient ($ > $ 15 years old) & \num{38425} \\
    		\# admissions where adult patient died 24 hours after the first admission & \num{35627} \\ \bottomrule
    	\end{tabular}
    }
	\label{tab:dataset_statistics}
\end{table}

\begin{table}[tb!]
    \renewcommand{\arraystretch}{1.2}
	\centering
	\caption{Baseline characteristics and in-hospital mortality outcome measures. Continuous variables are presented as \textit{Median [InterQuartile Range Q1-Q3]}; binary or categorical variables as \textit{Count (\%)}.}
	\label{tab:baseline_characteristics}
	\resizebox{\textwidth}{!}{%
			\begin{tabular}{@{}lllll@{}}
			\toprule
			&& \textbf{Overall} & \textbf{Dead at hospital} & \textbf{Alive at hospital} \\ \midrule
    		\textbf{General} &&& \\
            \# admissions && 35627 & 3738 & 31889 \\
			Age && 65.86 {[}52.72-77.97{]} & 73.85 {[}60.16-82.85{]} & 64.98 {[}52.04-77.21{]} \\
			Gender (female) && 15409 (43.24\%) & 1731 (46.31\%) & 13678 (42.88\%) \\
			First SAPS-II && 33.00 {[}25.00-42.00{]} & 48.00 {[}38.00-59.00{]} & 32.00 {[}24.00-40.00{]} \\
			First SOFA && 3.00 {[}2.00-6.00{]} & 6.00 {[}4.00-9.00{]} & 3.00 {[}2.00-5.00{]} \\ \midrule
			\textbf{Origin} &&&& \\
			Medical && 24720 (69.37\%) & 2969 (79.43\%) & 21751 (68.19\%) \\
			Emergency surgery && 6134 (17.21\%) & 663 (17.74\%) & 5471 (17.15\%) \\
			Scheduled surgery && 4783 (13.42\%) & 106 (2.84\%) & 4677 (14.66\%) \\ \midrule
			\textbf{Site} &&&& \\
			MICU && 12621 (35.42\%) & 1814 (48.53\%) & 10807 (33.88\%) \\
			MSICU && 5821 (16.33\%) & 691 (18.49\%) & 5130 (16.08\%) \\
			CCU && 5180 (14.54\%) & 523 (13.99\%) & 4657 (14.60\%) \\
			CSRU && 7264 (20.38\%) & 245 (6.55\%) & 7019 (22.00\%) \\
			TSICU && 4751 (13.33\%) & 465 (12.44\%) & 4286 (13.44\%) \\
			HR (bpm) && 84.00 {[}73.00-97.00{]} & 90.00 {[}75.00-107.00{]} & 84.00 {[}72.00-96.00{]} \\
			MAP (mmhg) && 76.00 {[}67.33-87.00{]} & 74.00 {[}64.67-86.00{]} & 77.00 {[}68.00-87.00{]} \\
			RR (cpm) && 18.00 {[}14.00-22.00{]} & 20.00 {[}16.00-24.00{]} & 18.00 {[}14.00-21.00{]} \\
			Na (mmol/l) && 138.00 {[}136.00-141.00{]} & 139.00 {[}135.00-142.00{]} & 138.00 {[}136.00-141.00{]} \\
			K (mmol/l) && 4.10 {[}3.80-4.60{]} & 4.20 {[}3.70-4.70{]} & 4.10 {[}3.80-4.60{]} \\
			$ HCO_3 $ (mmol/l) && 24.00 {[}21.00-26.00{]} & 22.00 {[}18.00-25.00{]} & 24.00 {[}21.00-26.00{]} \\
			WBC (\SI[per-mode=symbol]{e3}{\per\cubic\milli\meter}) && 11.00 {[}7.90-14.90{]} & 12.30 {[}8.00-17.20{]} & 10.80 {[}7.90-14.60{]} \\
			P/F ratio && 257.50 {[}180.00-352.50{]} & 218.66 {[}140.00-331.86{]} & 262.50 {[}187.00-355.00{]} \\
			Ht (\%) && 31.00 {[}26.00-36.00{]} & 31.00 {[}27.00-36.00{]} & 31.00 {[}26.00-36.00{]} \\
			Urea (mmol/l) && 1577.00 {[}968.00-2415.00{]} & 1020.00 {[}518.50-1780.00{]} & 1640.00 {[}1035.00-2470.00{]} \\
			Bilirubine (mg/dl) && 0.70 {[}0.40-1.70{]} & 1.00 {[}0.50-3.50{]} & 0.70 {[}0.40-1.50{]} \\
			Hospital LOS (days) && 7.08 {[}4.32-12.03{]} & 7.21 {[}3.31-14.44{]} & 7.07 {[}4.40-11.88{]} \\
			ICU death (\%) && 2860 (8.03\%) & 2860 (76.51\%) & -- \\
			Hospital death (\%) && 3738 (10.49\%) & -- & -- \\
            \bottomrule
		\end{tabular}
	}
\end{table} 

\subsection{Dataset Preprocessing}
\label{sec:mimiciii-preprocessing}
In this section, we describe in detail the cohort selection, data extraction, data cleaning and feature extraction methods we employed to preprocess our MIMIC-III dataset.

\subsubsection{Cohort Selection}
\label{sec:cohort_selection}
The first step of dataset preprocessing includes cohort selection. We used two sets of inclusion criterion to select the patients to prepare the benchmark datasets. First, we identified all the adult patients by using the age recorded at the time of ICU admission. Following previous studies~\citep{johnsonreproducibility}, in our work, all the patients whose age was \num{>15} years at the time of ICU admission is considered as an adult~\footnote{Note that in MIMIC III (v1.4), all the patients under the age of \num{15} years are referred to as \textit{neonates}.}. Second, for each patient, we only use their first admission in our benchmark datasets and for subsequent analysis, and dropped all their later admissions. This was done to prevent possible information leakage in the analysis, and to ensure similar experimental settings compared to the related works~\citep{johnsonreproducibility}.

\subsubsection{Data Extraction}
\label{sec:data_extraction}
There are 26 tables in the MIMIC-III (v1.4) relational database. Charted events such as laboratory tests, doctor notes  and fluids into/out of patients are stored in a series of 'events' tables. For the purpose of preparing benchmark datasets to predict clinical tasks, we extracted data for the selected cohort from the following tables: \textit{inputevents (inputevents\_cv}/\textit{inputevents\_mv)} (intake for patients monitored using Philips CareVue system/iMDSoft MetaVision system), \textit{outputevents} (output information for patients while in the ICU), \textit{chartevents} (all charted observations for patients), \textit{labevents} (laboratory measurements for patients both within the hospital and in outpatient clinics), and \textit{prescriptions} (medications ordered, and not necessarily administered, for a given patient). We selected these tables as they provide the most relevant clinical features for the prediction tasks considered in this work. We obtained the following two benchmark datasets:
\begin{itemize}
	\item MIMIC-III: This includes the data extracted from all the above tables for all the selected cohorts  in the entire MIMIC-III database.
	\item MIMIC-III (CareVue): This includes the data extracted from all the above tables for the selected cohorts who are included in the \textit{inputevents\_cv} table (\textit{inputevents} data recorded using Philips CareVue system) in the MIMIC-III database. MIMIC-III (CareVue) is a subset of MIMIC-III dataset and it roughly corresponds to the MIMIC-II~\citep{lee2011open} dataset.
\end{itemize}

\subsubsection{Data Cleaning}
\label{sec:dataset-cleaning}
The data extracted from MIMIC-III database has lots of erroneous entries due to noise, missing values, outliers, duplicate or incorrect records, clerical mistakes etc. We identified and handled the following three issues with the extracted data. First, we observed that there is inconsistency in the recording (units) of certain variables. For example, some of the prescriptions are recorded in `dose' and in `mg' units; while some variables in \textit{chartevents} and \textit{labevents} tables are recorded in both numeric and string data type. Second, some variables have multiple values recorded at the same time. Third, for some variables the observation was recorded as a range rather than a single measurement. We addressed these issues by these procedures:
\begin{itemize}
	\item To handle inconsistent units: We first obtain the percentage of each unit appearing in the database for a variable. If there is only one unit, we do nothing. For variables with multiple and inconsistent units, if a major unit accounts for \SI{\ge 90}{\percent} of the total number of records then we just keep all the records with the major unit and drop the other ones. For the rest of the variables/features which do not have a major unit, we convert all the units to a single unit based on accepted rules in literature~\footnote{\url{https://www.drugs.com/dosage/}} (For example: convert `mg' to `grams', `dose' to `ml' or `mg' based on the variable). We drop the features for which we cannot find correct rules for conversion.
	\item To handle multiple recordings at the same time: Depending on the variable, we either take the average or the summation of the multiple recordings present at the same time.
	\item To handle range of feature values: We take the median of the range to represent the value of the feature at a certain time point.
\end{itemize}

\subsubsection{Feature Selection and Extraction}
\label{sec:feature-selection}
We process the extracted benchmark datasets to obtain the features which will be used for the prediction tasks. To enable an exhaustive benchmarking comparison study, we select three sets of features as described below.

{
\small
\begin{longtable}[c]{p{0.18\textwidth}lll}
    \caption{Feature Set A: 17 features used in SAPS-II scoring system.} \\
	\toprule
	\textbf{Feature} & \textbf{Itemid} & \textbf{Name of Item} & \textbf{Table} \\ \midrule
    \endhead
	\rowcell{6}{glasgow coma scale} & 723 & GCSVerbal & chartevents \\
	& 454 & GCSMotor & chartevents \\
	& 184 & GCSEyes & chartevents \\
	& 223900 & Verbal Response & chartevents \\
	& 223901 & Motor Response & chartevents \\
	& 220739 & Eye Opening & chartevents \\ \midrule
	\rowcell{6}{systolic blood pressure} & 51 & Arterial BP {[}Systolic{]} & chartevents \\
	& 442 & Manual BP {[}Systolic{]} & chartevents \\
	& 455 & NBP {[}Systolic{]} & chartevents \\
	& 6701 & Arterial BP \#2 {[}Systolic{]} & chartevents \\
	& 220179 & Non Invasive Blood Pressure systolic & chartevents \\
	& 220050 & Arterial Blood Pressure systolic & chartevents \\ \midrule
	\rowcell{2}{heart rate} & 211 & Heart Rate & chartevents \\
	& 220045 & Heart Rate & chartevents \\ \midrule
	\rowcell{4}{body temperature} & 678 & Temperature F & chartevents \\
	& 223761 & Temperature Fahrenheit & chartevents \\
	& 676 & Temperature C & chartevents \\
	& 223762 & Temperature Celsius & chartevents \\ \midrule
	\rowcell{6}{pao2 / fio2 ratio} & 50821 & PO2 & labevents \\
	& 50816 & Oxygen & labevents \\
	& 223835 & Inspired O2 Fraction (FiO2) & chartevents \\
	& 3420 & FiO2 & chartevents \\
	& 3422 & FiO2 {[}Meas{]} & chartevents \\
	& 190 & FiO2 set & chartevents \\ \midrule
	\rowcell{26}{urine output} & 40055 & Urine Out Foley & outputevents \\
	& 43175 & Urine & outputevents \\
	& 40069 & Urine Out Void & outputevents \\
	& 40094 & Urine Out Condom Cath & outputevents \\
	& 40715 & Urine Out Suprapubic & outputevents \\
	& 40473 & Urine Out IleoConduit & outputevents \\
	& 40085 & Urine Out Incontinent & outputevents \\
	& 40057 & Urine Out Rt Nephrostomy & outputevents \\
	& 40056 & Urine Out Lt Nephrostomy & outputevents \\
	& 40405 & Urine Out Other & outputevents \\
	& 40428 & Orine Out Straight Cath & outputevents \\
	& 40086 & Urine Out Incontinent & outputevents \\
	& 40096 & Urine Out Ureteral Stent \#1 & outputevents \\
	& 40651 & Urine Out Ureteral Stent \#2 & outputevents \\
	& 226559 & Foley & outputevents \\
	& 226560 & Void & outputevents \\
	& 226561 & Condom Cath & outputevents \\
	& 226584 & Ileoconduit & outputevents \\
	& 226563 & Suprapubic & outputevents \\
	& 226564 & R Nephrostomy & outputevents \\
	& 226565 & L Nephrostomy & outputevents \\
	& 226567 & Straight Cath & outputevents \\
	& 226557 & R Ureteral Stent & outputevents \\
	& 226558 & L Ureteral Stent & outputevents \\
	& 227488 & GU Irrigant Volume In & outputevents \\
	& 227489 & GU Irrigant/Urine Volume Out & outputevents \\ \midrule
	serum urea nitrogen level & 51006 & Urea Nitrogen & labevents \\ \midrule
	\rowcell{2}{white blood cells count} & 51300 & WBC Count & labevents \\
	& 51301 & White Blood Cells & labevents \\ \midrule
	serum bicarbonate level & 50882 & BICARBONATE & labevents \\ \midrule
	\rowcell{2}{sodium level} & 950824 & Sodium Whole Blood & labevents \\
	& 50983 & Sodium & labevents \\ \midrule
	\rowcell{2}{potassium level} & 50822 & Potassium, whole blood & labevents \\
	& 50971 & Potassium & labevents \\ \midrule
	bilirubin level & 50885 & Bilirubin Total & labevents \\ \midrule
	\rowcell{2}{age} & \multirow{2}{*}{--} & intime & icustays \\
	&  & dob & patients \\ \midrule
	\textls[-20]{acquired immunodeficiency syndrome} & -- & icd9\_code & diagnoses\_icd \\ \midrule
	hematologic malignancy & -- & icd9\_code & diagnoses\_icd \\ \midrule
	metastatic cancer & -- & icd9\_code & diagnoses\_icd \\ \midrule
	\rowcell{2}{admission type} & \multirow{2}{*}{--} & curr\_service & services \\
	& & ADMISSION\_TYPE & admissions \\ \bottomrule
   	\label{tab:features_17}

\end{longtable}
}

\begin{itemize}
	\item \textbf{Feature Set A}: This feature set consists of the 17 features used in the calculation of the SAPS-II score~\citep{le1993new}.
	For these features, we drop outliers in the data according to medical knowledge and merge relevant features. For example, for the Glasgow Coma Scale score denoted as GCS score, we sum the GCSVerbal, GCSMotor and GCSEyes values; for the urine output, we sum the features representing urine; and for body temperature, we convert Fahrenheit to Celsius scale. Note that the SAPS II score features are hand-chosen and processed, and thus, we refer to them as `Processed' features instead of `raw' features. Table~\ref{tab:features_17} lists all the 17 processed features and their corresponding entries in the MIMIC-III database table. In our experiments, some of these features such as chronic diseases, admission type and age are treated as non-time series features, and the remaining features are treated as time series features.
		
	\item \textbf{Feature Set B}: This feature set consists of the 20 features related to the 17 features used in SAPS-II score. Instead of preprocessing the 17 features as done to obtain Feature set A, here we consider all the raw values of the 17 SAPS-II score features. In particular,  we do not remove outliers and we only drop values below 0. For the GCS score, we treat GCSVerbal, GCSMotor and GCSEyes as separate features. We also consider PaO2 and FiO2 as individual features instead of calculating the PF-ratio (PaO2/FiO2 ratio). This feature set was built to study how the prediction models perform on the `raw' clinical features.
	
	\item \textbf{Feature Set C}: This feature set consists of 135 raw features selected from the 5 tables mentioned in section~\ref{sec:data_extraction} and includes the 20 features of Feature set B. These 135 features were chosen based on their low missing rate, from more than \num{20000} features available in the 5 tables mentioned in section~\ref{sec:data_extraction}. Similar to feature set B, we did not preprocess this dataset (i.e. did not apply hand-crafted processing rules) and used the raw values of the features. It is worth noting that a few features appear multiple times as they were present in multiple tables. For example, Glucose appears in both Chart and Lab events, and was included in the feature set. This feature set was selected to study if the prediction models can automatically learn feature representations from a large number of raw clinical time series data and at the same time obtain better results on the prediction tasks. Table~\ref{tab:features_100} in the Appendix lists all the features of this feature set C. 	
\end{itemize}

We extract the above three feature sets from MIMIC-III and MIMIC-III (CareVue) datasets.
After the feature selection, we obtained the non-time series and time series features which will be used in the experiments.
We extracted the features from first 24 hours and first 48 hours after admission to ICU. Each time series feature is sampled every 1 hour. To fill-in missing values, we performed forward and backward imputation. For some patients, certain features might be completely missing. We performed mean imputation for these cases during the training and validation stage of the experiments. We obtain summary statistics of time-series features for models which are not capable of handling temporal data.

	\section{Benchmarking Experiments}
	\label{sec:experiments}
	In this section, we describe in detail the benchmark prediction tasks, the prediction algorithms and their implementation, and report the experimental results. 

\subsection{Benchmark Prediction Tasks}
\label{sec:exp-tasks}
Here, we describe the benchmark prediction tasks which represent some of the important problems in critical care research. They have been well-studied in the medical community~\citep{knaus1981apache,silva2012predicting,kim2011comparison}, and these tasks have been commonly used to benchmark machine learning algorithms~\citep{pirracchio2016mortality, harutyunyan2017multitask}.

\subsubsection{Mortality Prediction}
\label{sec:mortality_prediction}
Mortality prediction is one of the primary outcomes of interest of an hospital admission.
We formulate mortality as a binary classification task, where the label indicates the death event for a patient. We define the following mortality prediction benchmark tasks:
\begin{itemize}
	\item In-hospital mortality prediction: Predict whether the patient dies during the hospital stay after admitted to an ICU.
	\item Short-term mortality prediction: Predict whether the death happens within a short duration of time after the patient is admitted to the ICU. For this task, we define the 2-day and 3-day mortality prediction tasks where the patient dies within 2-days and 3-days respectively after admitted to ICU. For first 24-hour data, we can predict 2-day and 3-day mortality, while for the first 48-hour data we only predict 3-day mortality.
	\item Long-term mortality prediction: This task involves predicting if the patient dies after a long time since being discharged from the hospital. For this task, we consider the 30-day and 1-year mortality prediction tasks where the patient dies within 30-days or 1 year after being discharged from the hospital. Note that we still use only the first 24-hour data and first 48-hour data to predict 30-days and 1-year mortality.
\end{itemize}

Table~\ref{tab:mortality_label} shows the mortality label statistics of the entire MIMIC-III dataset. The details about how the mortality labels are obtained from MIMIC-III database is explained in the Appendix.

\begin{table}[htb]
    \renewcommand{\arraystretch}{1.2}
	\centering
	\caption{Label statistics of mortality prediction task.}
	\label{tab:mortality_label}
	\resizebox{\textwidth}{!}{%
		\begin{tabular}{l*5{S[table-format=1.3]}S[table-format=5]}
			\toprule
			{\multirow{2}{*}{\tabincell{c}{\textbf{MIMIC-III} \\ \textbf{Datasource}}}} & \multicolumn{5}{c}{\textbf{Mortality label ratio w.r.t total admissions}} & {\multirow{2}{*}{\textbf{\# Admissions}}} \\ \cmidrule{2-6}
			 & {\textbf{In-hospital}} & {\textbf{2-day}} & {\textbf{3-day}} & {\textbf{30-day}} & {\textbf{1-year}} &  \\ \midrule
			Metavision (2008-2012) & 0.096 & 0.015 & 0.014 & 0.124 & 0.232 & 15376 \\
			CareVue (2001-2008) & 0.111 & 0.014 & 0.015 & 0.134 & 0.261 & 20261 \\
			All sources (2001-2012) & 0.105 & 0.014 & 0.015 & 0.129 & 0.248 & 35637 \\
            \bottomrule
		\end{tabular}
	}
\end{table}

\subsubsection{ICD-9 Code Group Prediction}
\label{sec:icd9_code_prediction}
In this benchmarking task, we predict the ICD-9 diagnosis code group (e.g. respiratory system diagnosis) for each admission. ICD (stands for International Statistical Classification of Diseases and Related Health Problems) codes are used to classify diseases and a wide variety of symptoms, signs, causes of injury or disease, etc. Nearly every health condition can be assigned an unique ICD-9 code group where each group usually include a set of similar diseases.
In our work, we group all the ICD-9 codes for an ICU admission into 20 diagnosis groups\footnote{\url{http://tdrdata.com/ipd/ipd_SearchForICD9CodesAndDescriptions.aspx}} and treat this task as a multi-task prediction problem.
Table~\ref{tab:icd9_label} shows the ICD-9 code group label statistics of the MIMIC-III dataset.

\begin{table}[htb]
    \renewcommand{\arraystretch}{1.2}
	\centering
	\caption{ICD-9 code group label statistics. For each ICD-9 code group, the entry denotes the ratio of number of patients who have been assigned that ICD-9 code to the total number of patients in the dataset.}
	\label{tab:icd9_label}
    \small
	\begin{tabular}{cc*3S[table-format=1.3]}
		\toprule
        {\tabincell{c}{\textbf{ICD-9 Code} \\ \textbf{Group}}} & {\tabincell{c}{\textbf{ICD-9 Code} \\ \textbf{Range}}} & {\tabincell{c}{\textbf{MIMIC-III} \\ \textbf{Metavision} \\ (2008-2012)}} & {\tabincell{c}{\textbf{MIMIC-III} \\ \textbf{CareVue} \\ (2001-2008)}} & {\tabincell{c}{\textbf{MIMIC-III} \\ \textbf{All Sources} \\ (2001-2012)}} \\ \midrule
		1       & 001 - 139            & 0.302               & 0.225            & 0.258        \\
		2       & 140 - 239            & 0.201               & 0.151            & 0.172        \\
		3       & 240 - 279            & 0.765               & 0.629            & 0.688        \\
		4       & 280 - 289            & 0.445               & 0.311            & 0.369        \\
		5       & 290 - 319            & 0.416               & 0.244            & 0.318        \\
		6       & 320 - 389            & 0.424               & 0.195            & 0.294        \\
		7       & 390 - 459            & 0.846               & 0.820            & 0.831        \\
		8       & 460 - 519            & 0.504               & 0.468            & 0.484        \\
		9       & 520 - 579            & 0.461               & 0.339            & 0.391        \\
		10      & 580 - 629            & 0.461               & 0.352            & 0.399        \\
		11      & 630 - 679            & 0.003               & 0.005            & 0.004        \\
		12      & 680 - 709            & 0.119               & 0.090            & 0.102        \\
		13      & 710 - 739            & 0.266               & 0.133            & 0.190        \\
		14      & 740 - 759            & 0.042               & 0.032            & 0.036        \\
		15      & 780 - 789            & 0.411               & 0.251            & 0.320        \\
		16      & 790 - 796            & 0.115               & 0.064            & 0.086        \\
		17      & 797 - 799            & 0.050               & 0.016            & 0.030        \\
		18      & 800 - 999            & 0.453               & 0.448            & 0.450        \\
		19      & V Codes            & 0.634               & 0.362            & 0.479        \\
		20      & E Codes            & 0.429               & 0.263            & 0.335        \\
        \bottomrule
	\end{tabular}
\end{table}

\subsubsection{Length of Stay Prediction}
\label{sec:length_of_stay_prediction}
\begin{figure}[htb]
	\centering
    \subfigure{
	   \includegraphics[width=0.98\columnwidth]{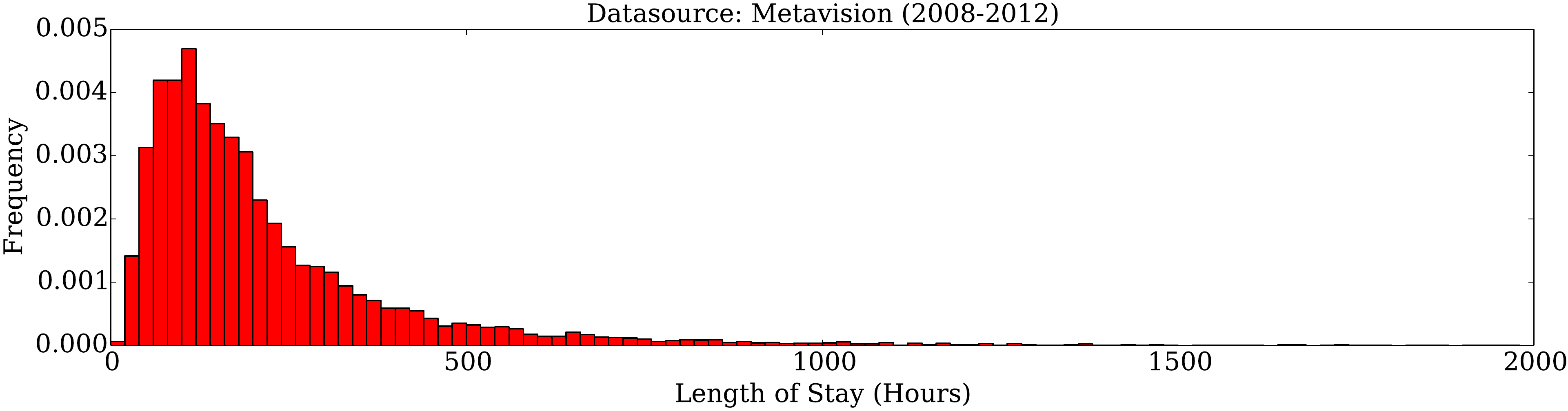}
    }
    \subfigure{
	   \includegraphics[width=0.98\columnwidth]{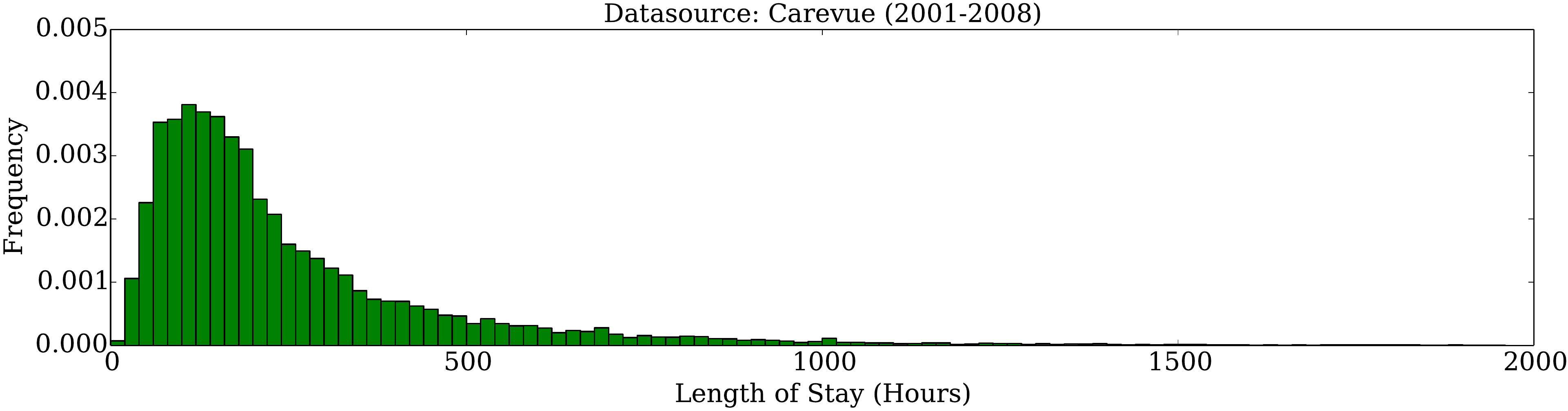}
    }
    \subfigure{
	   \includegraphics[width=0.98\columnwidth]{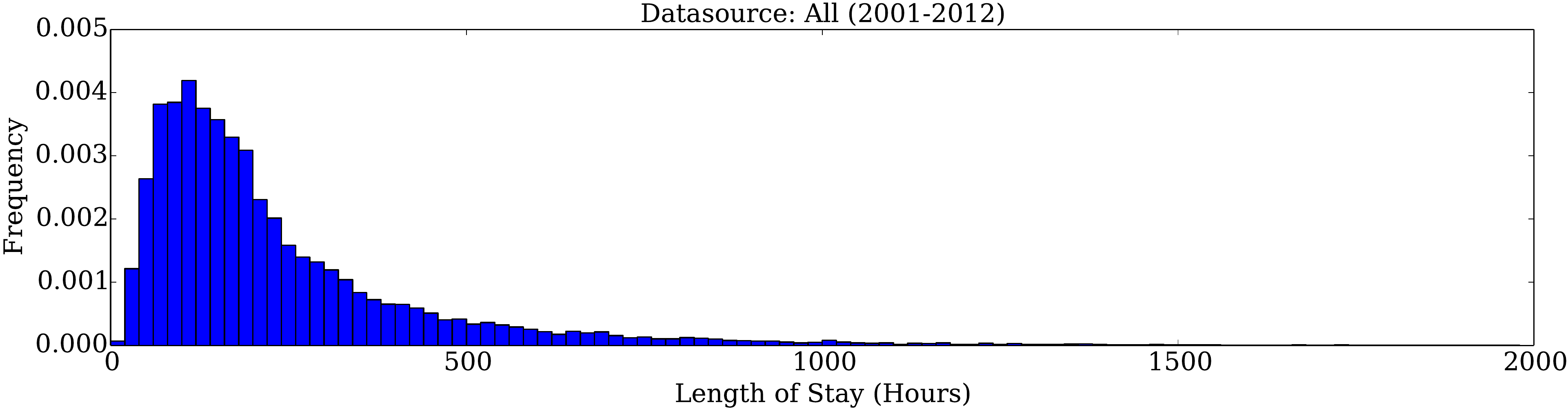}
    }
	\caption{Distribution of length of stay. Values above 2000 hours are not shown in the figure.}
	\label{fig:los_stat}
\end{figure}

In this benchmarking task, we predict the length of stay for each admission. We define the length of stay of an admission as total duration of hospital stay, i.e. the length of time interval between hospital admission and discharge from the hospital. We treat length of stay prediction task as a regression problem.
Figure~\ref{fig:los_stat} shows the distribution of length of stay of the MIMIC-III benchmark datasets.

\subsection{Prediction Algorithms}
\label{sec:exp-methods}
In this section, we describe all the prediction algorithms and the scoring systems that we have used for benchmarking tasks on MIMIC-III datasets.

\subsubsection{Scoring Methods}
\label{sec:score_models}
\paragraph{SAPS-II}
SAPS-II~\citep{le1993new} stands for Simplified Acute Physiology Score and
it is a ICU scoring system designed to measure the severity of the disease for patients admitted to an ICU. A point score is calculated for each of the 12 physiological features mentioned in Table~\ref{tab:features_17} and a final SAPS-II score $S$ is obtained as the sum of all point scores. Note that SAPS-II score is calculated using the data collected within the first 24 hours of an ICU admission. After SAPS-II score is obtained, the individual mortality prediction can be calculated as \citep{pirracchio2016mortality}:
\begin{equation*} \log\dfrac{p_{\mathrm{death}}}{1-p_{\mathrm{death}}}=-7.7631+0.0737 \cdot S+0.9971\cdot \log\left(1+S\right)
\end{equation*}

\paragraph{SOFA}
SOFA~\citep{vincent1996sofa} is the Sepsis-related Organ Failure Assessment score (also referred to as the Sequential Organ Failure Assessment score) and it is used to describe organ dysfunction/failure of a patient in the ICU. The mortality prediction based on SOFA can be obtained by regressing the mortality on the SOFA score using a main-term logistic regression model.

\paragraph{New SAPS-II}
A new SAPS-II scoring method was defined by \citet{pirracchio2016mortality}. It is a modified version of SAPS-II and is obtained by fitting a main-term logistic regression model using the same explanatory variables as those used in the original SAPS-II score calculation.

\subsubsection{Super Learner Models}
\label{sec:superlearner}
Super Learner~\citep{van2007super, polley2010super} is a supervised learning algorithm that is designed to find the optimal combination from a set of prediction algorithms. It represents an asymptotically optimal learning system and is built on the theory of cross-validation. This algorithm requires a collection of user-defined machine learning algorithms such as logistic regression, regression trees, additive models, (shallow) neural networks, and random forest. The algorithm then estimates the
risk associated to each algorithm in the provided collection using cross-validation.
One round of cross-validation involves partitioning a sample of data into complementary
subsets, performing the analysis on one subset (called the training set), and
validating the analysis on the other subset (called the validation set or testing set). To
reduce variability, multiple rounds of cross-validation are performed using different
partitions, and the validation results are averaged over the rounds. From this estimation
of the risk associated with each candidate algorithm, the Super Learner builds
an aggregate algorithm obtained as the optimal weighted combination of the candidate
algorithms. Table~\ref{table:algo_sl} shows the algorithms used in the Super Learner
algorithm~\citep{pirracchio2016mortality} and their implementation available in the R and Python languages.
In the experiments section, we will compare and discuss the results of Super Learner using these programming languages. Following \citet{pirracchio2016mortality}, we consider two variants of Super Learner algorithm, namely, Super Learner I: Super Learner with categorized variables, and Super Learner II: Super Learner with non-transformed variables. Note that Super Learner-I is applicable only for Feature set A, while Super Learner-II algorithm can be used with all the Feature sets A, B and C.

\begin{table}[htb]
	\centering
	\caption{Algorithms used in Super Learner with corresponding R packages and Python libraries.}
	\label{table:algo_sl}
	\resizebox{\textwidth}{!}{
    \renewcommand{\arraystretch}{1.2}
		\begin{tabular}{@{}llll@{}}
			\toprule
			\textbf{Algorithm} &  & \textbf{R packages} & \textbf{Python libraries}\\
			\midrule
			Standard logistic regression &  & SL.glm & sklearn.linear\_model.LogisticRegression\\\hline
			\tabincell{l}{Logistic regression \\
				based on the AIC} &  & SL.stepAIC & sklearn.linear\_model.LassoLarsIC \\\hline
			Generalized additive model &  & SL.gam & \tabincell{l}{pygam.LinearGAM \\pygam.LogisticGAM} \\\hline
			\tabincell{l}{Generalized linear model\\ with penalized maximum likelihood} &  & SL.glmnet & sklearn.linear\_model.ElasticNet \\\hline
			\tabincell{l}{Multivariate adaptive polynomial \\ spline regression} &  & SL.polymars & pyearth.Earth\\\hline
			Bayesian generalized linear model &  & SL.bayesglm & sklearn.linear\_model.BayesianRidge  \\\hline
			\tabincell{l}{Generalized boosted \\ regression model} & & SL.gbm & \tabincell{l}{sklearn.ensemble.GradientBoostingRegressor \\ sklearn.ensemble.GradientBoostingClassifier} \\\hline
			Neural network &  & \tabincell{l}{SL.nnet } & \tabincell{l}{sklearn.neural\_network.MLPRegressor \\ sklearn.neural\_network.MLPClassifier} \\\hline
			Bagging classification trees &  & \tabincell{l}{SL.ipredbagg} & \tabincell{l}{sklearn.ensemble.BaggingRegressor \\ sklearn.ensemble.BaggingClassifier}\\\hline
			\tabincell{l}{Pruned recursive partitioning \\ and Regression Trees} &  & SL.rpartPrune & -- \\\hline
			Random forest &  & SL.randomForest & \tabincell{l}{sklearn.ensemble.RandomForestRegressor \\ sklearn.ensemble.RandomForestClassifier} \\\hline
			Bayesian additive regression trees &  & SL.bartMachine  & -- \\
			\bottomrule
		\end{tabular}
    }
	\end{table}

\subsubsection{Deep Learning Models}
Deep Learning Models (also called Deep Neural Networks or Deep models)~\citep{lecun2015deep} have become a successful approach for automated extraction of complex data representations for end-to-end training. Deep models consist of a layered, hierarchical architectures of neurons for learning and representing data. The hierarchical learning architecture is motivated by artificial intelligence emulating the deep, layered learning process of the primary sensorial areas of the neocortex in the human brain, which automatically extracts features and abstractions from the underlying data~\citep{larochelle2009exploring,bengio2013representation}.
In a deep learning model, each neuron receives one or more inputs and sums them to produce an output (or activation). Each neuron in the hidden layers is assigned a weight that is considered for the outcome classification, but this weight is itself learned from its previous layers. The hidden layers thus can use multidimensional input data and introduce progressively non-linear weight combinations to the learning algorithm.

The main advantage of the deep learning approach is its ability to automatically learn good feature representations from raw data, and thus significantly reduce the effort of handcrafted feature engineering. In addition, deep models learn distributed representations of data, which enables generalization to new combinations of the values of learned features beyond those seen during the training process. Deep Learning models have yielded outstanding results in several applications, including speech recognition~\citep{dahl2010phone,dahl2012context}, computer vision~\citep{krizhevsky2012imagenet,jia2014caffe,szegedy2015going}, and natural language processing ~\citep{mikolov2011empirical,bordes2012joint,mikolov2013linguistic,mikolov2013distributed}. Recent research has shown that deep learning methods achieve state-of-the-art performance in analyzing health-related data, such as ICU mortality prediction~\citep{johnsonreproducibility}, phenotype discovery~\citep{che2015deep} and disease prediction~\citep{che2015distilling}. These works have demonstrated the strong performance by deep learning models in health care applications, which significantly alleviates the tedious work on feature engineering and extraction. Here, we will first
briefly introduce two types of deep models namely Feedforward
neural networks (FFN), which is a standard neural network
structure, and Recurrent Neural Networks (RNN) which is used for modeling sequence and time series data. After that, we will describe our proposed Multi-modal deep learning model, a combination of FFN and RNN, which will be used in the benchmarking experiments.

\label{sec:DL_models}
\paragraph{Feedforward Neural Networks}
A multilayer feedforward network~\citep{hornik1989multilayer} (FFN) is a neural network with multiple nonlinear layers and possibly one prediction layer on the top to solve classification task.
The first layer takes $\vct{X}$ as the input, and the output of each layer is used as the input of the next layer. The transformation of each layer $l$ can be written as
\begin{align*}
\vct{X}^{(l+1)}= f^{(l)}(\vct{X}^{(l)}) = s^{(l)}\left( \mat{W}^{(l)}\vct{X}^{(l)} + \vct{b}^{(l)} \right)
\end{align*}
where $\mat{W}^{(l)}$ and $\vct{b}^{(l)}$ are respectively the weight matrix and bias vector of layer $l$, and $s^{(l)}$ is a nonlinear activation function, which usually is a \textit{logistic sigmoid}, \textit{tanh}, or \textit{ReLU}~\citep{nair2010rectified} function.
We optimize the cross-entropy prediction loss and get the prediction output from the topmost prediction layer.

\begin{figure}[htb]
    \centering
    \hspace{0.05\columnwidth}
    \subfigure[\label{fig:GRU}Gated recurrent unit.]{
        \includegraphics[width=0.35\columnwidth]{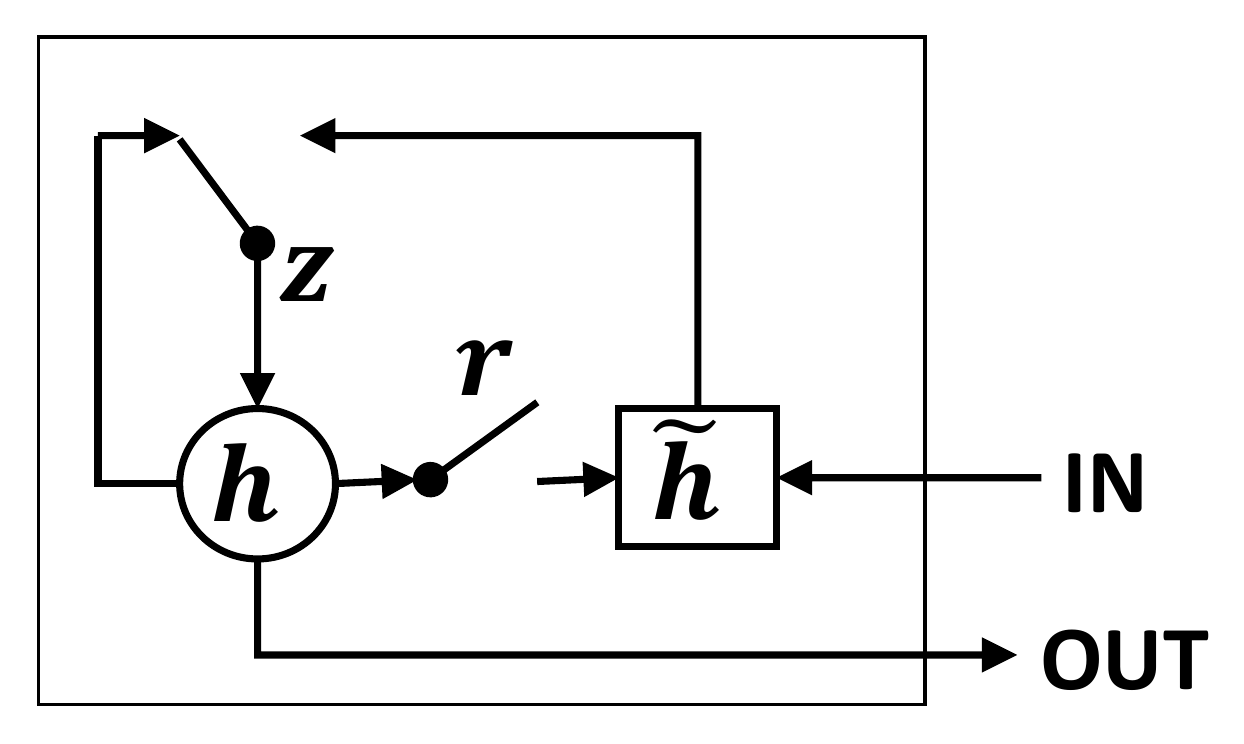}
    }
    \hfill
    \subfigure[\label{fig:MMDL} Multimodal deep learning models.]{
    	\includegraphics[width=0.35\columnwidth]{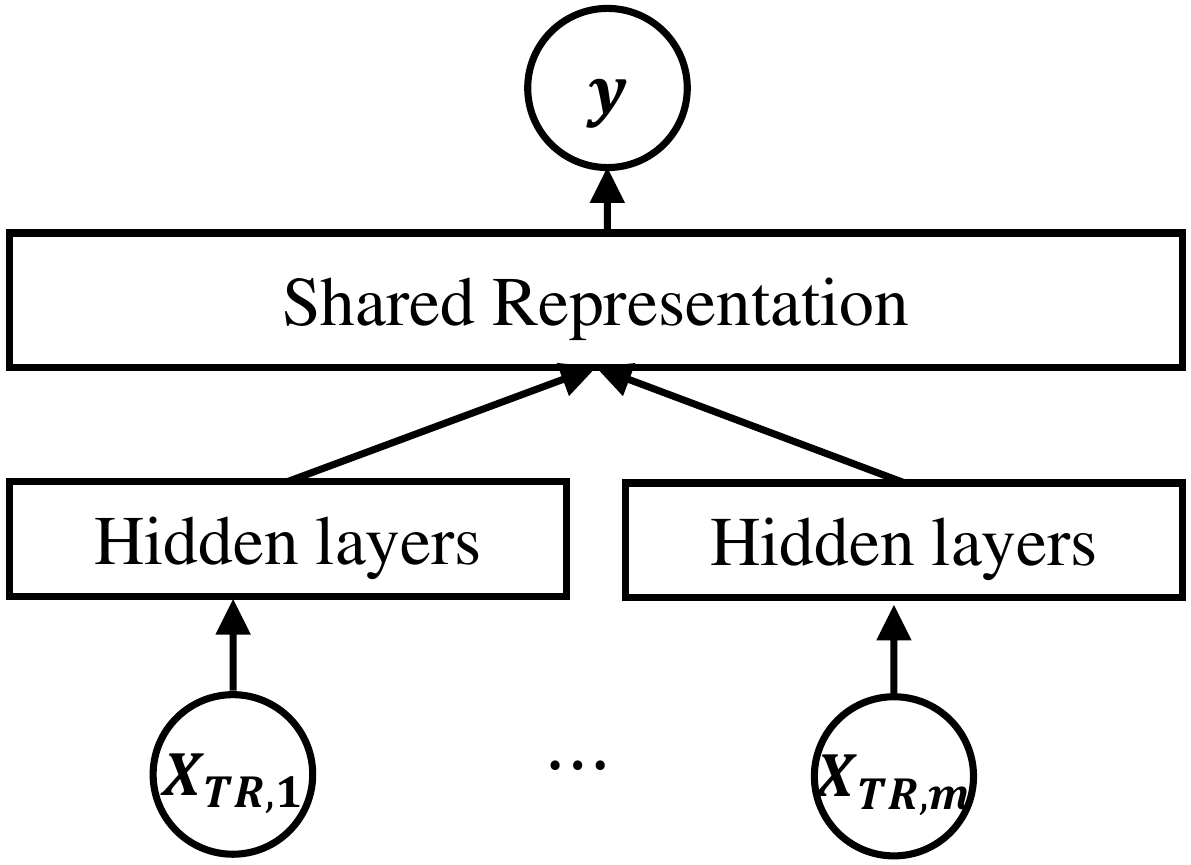}
    }
    \hspace{0.05\columnwidth}
    \caption{Deep learning models. In multimodal deep models, $ X_{(.)} $ represents the different inputs including temporal and non-temporal features, and $ y $ is the output.}
\end{figure}

\paragraph{Recurrent Neural Networks}
Recurrent neural network (RNN) models have been shown to be successful at modeling sequences and time series data~\citep{williams1989learning}. RNN with simple activations are incapable of capturing long term dependencies, and hence their variants such as Long Short Term Memory (LSTM)~\citep{hochreiter1997long} and Gated Recurrent Unit (GRU)~\citep{cho2014properties} have become popular due to their ability to capture long-term dependencies using memory and gating units.
GRU can be considered as a simplified version of LSTM and it has been shown that GRU has similar performance compared to LSTM~\citep{chung2014empirical}.
The structure of GRU is shown in Figure~\ref{fig:GRU}. Let $\vct{x}_t \in \R^P$ denotes the variables at time $t$, where $ 1\le t \le T$. At each time $t$, GRU has a reset gate $r_t^j$ and an update gate $z_t^j$ for each of the hidden state $h_t^j$.
The update function of GRU is shown as follows:
\begin{align*}
\vct{z}_t &= \sigma \left( \mat{W}_z \vct{x}_t + \mat{U}_z \vct{h}_{t-1} + \vct{b}_z \right)
& \vct{r}_t &= \sigma \left( \mat{W}_r \vct{x}_t + \mat{U}_r \vct{h}_{t-1} + \vct{b}_r \right) \\
\tilde{\vct{h}}_t &= tanh \left( \mat{W} \vct{x}_t + \mat{U} (\vct{r}_t \odot \vct{h}_{t-1})  + \vct{b} \right)
& \vct{h}_t &= (\vct{1}-\vct{z}_t) \odot \vct{h}_{t-1} + \vct{z}_t \odot \tilde{\vct{h}}_t
\end{align*}
where matrices $\mat{W}_z,\mat{W}_r,\mat{W}, \mat{U}_z,\mat{U}_r,\mat{U}$ and vectors $\vct{b}_z, \vct{b}_r, \vct{b}$ are model parameters.

\paragraph{Multimodal Deep Learning Model (MMDL)}
As our benchmarking datasets come from multiple tables and includes both temporal and non-temporal data, multimodal deep learning models~\citep{srivastava2012multimodal} can be used to shared learn representations for the prediction tasks. Here, we propose a deep learning framework called as Multimodal Deep Learning Model (MMDL) to learn shared representations from multiple modalities using an ensemble of FFN and GRU deep learning models. The key idea is to use a shared representation layer to capture the correlations of modalities or to learn a similarity of modalities in representation space, which is beneficial when limited data is available from multiple modalities. Data from each of
the different tables can be treated as a separate modality. For simplicity, in MMDL, we treat all the temporal features as one modality and all non-temporal features as another modality. Figure~\ref{fig:MMDL} shows an illustration of our MMDL framework with a common layer to learn the shared representations of modalities. MMDL uses FFN and GRU to handle non-temporal and temporal features respectively, and learns their shared latent representations for prediction tasks.

\subsection{Implementation Details}
\label{sec:exp-implementation}

We implement the Super Learner algorithm using R packages and Python libraries listed in Table~\ref{table:algo_sl}. The deep learning models are implemented in Theano~\citep{bastien2012theano} and Keras~\citep{chollet2015keras} platforms. For all the prediction methods, we conduct a five-fold cross validation and report the mean and standard error of performance scores. We use Area under the ROC curve (AUROC) and Area under Precision-Recall Curve (AUPRC) as the evaluation metrics to report the prediction model's performance on classification tasks, and use Mean Squared Error (MSE) to report results on the regression task.

All the deep learning models are trained with RMSProp optimizer method with learning rate of \num{0.001} on classification tasks and \num{0.005} on regression tasks. The batch size is chosen as \num{100} and the max epoch number is fixed at \num{250}. Early stopping with best weight and batch normalization are applied during training. MMDL is a combination of FFN and GRU, in which FFN part handles non-temporal features and GRU part handles temporal features. The structure of MMDL used in our experiments is shown in Figure~\ref{fig:ffn_lstm_struct} in the Appendix.

All the data is divided to 5 folds with stratified cross validation, and standardization is done to the whole dataset with the mean and standard error of the training set.
To enable reproducibility of our results, we will be releasing our preprocessing codes and benchmark prediction task codes on the Github soon.

\subsection{Results}
\label{sec:exp-results}

In this section, we report the benchmarking results of all the prediction algorithms on the MIMIC-III datasets. We answer the following questions: (a) How do the Deep Learning models compare to the Super Learner algorithm and scoring systems? (b) What is the performance of prediction methods on the different feature sets?

\subsubsection{Performance of Super Learner Algorithm Implementations}
First, we compare the performance of Super Learner-R and Super Learner-Python softwares on in-hospital mortality prediction task using feature set A i.e. 17 processed features collected in the first 24 hours of ICU admission from MIMIC-III dataset. The result in Table~\ref{table:r_py_eval} shows that Super Learner-Python performs slightly better than Super Leaner-R implementation. Moreover, Super Learner-Python can be evaluated significantly faster than Super Learner-R. Thus, in the following experiments, we will only report the results of Super Learner-Python version (unless otherwise stated) to evaluate and benchmark Super Learner algorithm on different tasks.

\begin{table}[h]
    \renewcommand{\arraystretch}{1.2}
	\centering
	\caption{Comparison of Super Learner-R and Super Learner-Python software versions on in-hospital mortality prediction task using Feature set A extracted from the first 24-hour data of MIMIC-III. Running time refers to total time taken to perform cross-validation evaluation.}
	\label{table:r_py_eval}
	\resizebox{\textwidth}{!}{%
	\begin{tabular}{ll*2{S[table-format=1.4] @{$\ \pm\ $} S[table-format=1.4]} l}
		\toprule
		&                & \multicolumn{2}{c}{\textbf{AUROC Score}}                   & \multicolumn{2}{c}{\textbf{AUPRC Score}}                    & \textbf{Running Time}\\ \midrule
		\multirow{2}{*}{SuperLearner-I}  & R version      & 0.8402 & 0.0021          & 0.4304 & 0.0130          & 36 hours \\
		& Python version &  \bfseries 0.8448 & \bfseries 0.0038 &  \bfseries 0.4351 & \bfseries 0.0139 & 30 minutes\\ \midrule
		\multirow{2}{*}{SuperLearner-II} & R version      & 0.8646 & 0.0023          & 0.4917 & 0.0093          & 28 hours\\
		& Python version &  \bfseries 0.8701 & \bfseries 0.0053 &  \bfseries 0.4991 & \bfseries 0.0107 & 25 minutes\\ \bottomrule
	\end{tabular}
}
\end{table}

\subsubsection{Mortality Prediction Task Evaluation}
\label{sec:results-mortality}
Here, we report the performance of all methods described in Section~\ref{sec:exp-methods} on the mortality prediction tasks for benchmark datasets MIMIC-III and MIMIC-III (CareVue). We report the mean and standard deviation of AUROC and AUPRC for all the tasks.

\paragraph{In-hospital Mortality Prediction}

Tables~\ref{table:mimic-17-mor0} and \ref{table:mimic-17-mor1} show the in-hospital mortality prediction task results of all the prediction algorithms on Feature Set A of MIMIC-III and MIMIC-III (CareVue) datasets for both 24 hour and 48 hour data. From these tables, we observe that deep learning models such as MMDL and RNN perform better than all the other models on 48-hour data. On 24-hour data, we observe that Super Learner II model obtains slightly better results than deep learning model.

\begin{table}[b]
    \renewcommand{\arraystretch}{1.2}
	\centering
	\caption{In-hospital mortality task on MIMIC-III using feature set A.}
	\label{table:mimic-17-mor0}
	\resizebox{\textwidth}{!}{
\begin{tabular}{@{}ll *2{l*2{S[table-format=1.4]@{$\ \pm\ $}S[table-format=1.4]}} @{}}
			\toprule
			\multirow{2}{*}{\textbf{Method}} 	&  \multirow{2}{*}{\textbf{Algorithm}} 	&&  \multicolumn{4}{c}{\textbf{Feature Set A, 24-hour data}} 	&&  \multicolumn{4}{c}{\textbf{Feature Set A, 48-hour data}} \\ \cmidrule{4-7} \cmidrule{9-12}
			& &	&  \multicolumn{2}{c}{AUROC Score} 	&  \multicolumn{2}{c}{AUPRC Score} 	&&  \multicolumn{2}{c}{AUROC Score}	&  \multicolumn{2}{c}{AUPRC Score}  \\ \midrule
			\multirow{3}{*}{Score Methods} 	&  SAPS-II 	&&  0.8035 	&  0.0044 	&  0.3586 	&  0.0052 	&&  0.8046 	&  0.0083 	&  0.3373 	&  0.0141  \\
			& New SAPS-II 	&&  0.8235 	&  0.0042 	&  0.3989 	&  0.0120 	&&  0.8252 	&  0.0036 	&  0.3823 	&  0.0119  \\
			& SOFA 	&&  0.7322 	&  0.0038 	&  0.3191 	&  0.0085 	&&  0.7347 	&  0.0094 	&  0.2852 	&  0.0167  \\ \midrule
			\multirow{12}{*}{Super Learner} 	&  SL.glm 	&&  0.8235 	&  0.0042 	&  0.3987 	&  0.0120 	&&  0.8251 	&  0.0037 	&  0.3828 	&  0.0112  \\
			& SL.gbm 	&&  0.8435 	&  0.0034 	&  0.4320 	&  0.0125 	&&  0.8452 	&  0.0052 	&  0.4163 	&  0.0121  \\
			& SL.nnet 	&&  0.8388 	&  0.0044 	&  0.4200 	&  0.0135 	&&  0.8381 	&  0.0055 	&  0.3989 	&  0.0131  \\
			& SL.ipredbagg 	&&  0.7556 	&  0.0064 	&  0.3104 	&  0.0084 	&&  0.7510 	&  0.0078 	&  0.2811 	&  0.0121  \\
			& SL.randomforest 	&&  0.7576 	&  0.0085 	&  0.3104 	&  0.0084 	&&  0.7538 	&  0.0095 	&  0.2830 	&  0.0121  \\
			& SuperLearner-I 	&&  0.8448 	&  0.0038 	&  0.4351 	&  0.0139 	&&  0.8465 	&  0.0057 	&  0.4190 	&  0.0124  \\ \cmidrule{2-12}
			& SL.glm 	&&  0.8024 	&  0.0043 	&  0.3804 	&  0.0043 	&&  0.8013 	&  0.0021 	&  0.3559 	&  0.0238  \\
			& SL.gbm 	&&  0.8628 	&  0.0037 	&  0.4840 	&  0.0078 	&&  0.8518 	&  0.0049 	&  0.4259 	&  0.0209  \\
			& SL.nnet 	&&  0.8490 	&  0.0079 	&  0.4587 	&  0.0058 	&&  0.8383 	&  0.0058 	&  0.4028 	&  0.0180  \\
			& SL.ipredbagg 	&&  0.8060 	&  0.0069 	&  0.4087 	&  0.0110 	&&  0.7816 	&  0.0028 	&  0.3455 	&  0.0159  \\
			& SL.randomforest 	&&  0.7977 	&  0.0079 	&  0.3958 	&  0.0124 	&&  0.7813 	&  0.0059 	&  0.3496 	&  0.0200  \\
			& SuperLearner-II 	&&  \bfseries 0.8673 	&  \bfseries 0.0045 	&  \bfseries 0.4968 	&  \bfseries 0.0097 	&&  0.8595 	&  0.0035 	&  0.4422 	&  0.0200  \\ \midrule
			\multirow{3}{*}{Deep Learning} 	&  FFN 	&&  0.8496 	&  0.0047 	&  0.4632 	&  0.0074 	&&  0.8375 	&  0.0041 	&  0.4090 	&  0.0169  \\
            & RNN  	&& 0.8544	& 0.0053	& 0.4519	& 0.0145	&& 0.8618	& 0.0059	& 0.4458	& 0.0144  \\
			& MMDL 	&&  0.8664 	&  0.0056 	&  0.4776 	&  0.0162 	&&  \bfseries 0.8737 	&  \bfseries 0.0045 	&  \bfseries 0.4714 	&  \bfseries 0.0176  \\ \bottomrule
		\end{tabular}
	}
\end{table}

\begin{table}[b]
    \renewcommand{\arraystretch}{1.2}
	\centering
	\caption{In-hospital mortality task on MIMIC-III (Carvue) using feature set A.}
	\label{table:mimic-17-mor1}
	\resizebox{\textwidth}{!}{
		\begin{tabular}{@{}ll *2{l*2{S[table-format=1.4]@{$\ \pm\ $}S[table-format=1.4]}} @{}}
			\toprule
			\multirow{2}{*}{\textbf{Method}} 	&  \multirow{2}{*}{\textbf{Algorithm}} 	&&  \multicolumn{4}{c}{\textbf{Feature Set A, 24-hour data}} 	&&  \multicolumn{4}{c}{\textbf{Feature Set A, 48-hour data}} \\ \cmidrule{4-7} \cmidrule{9-12}
			&	&	&  \multicolumn{2}{c}{AUROC Score} 	&  \multicolumn{2}{c}{AUPRC Score} 	&&  \multicolumn{2}{c}{AUROC Score}	&  \multicolumn{2}{c}{AUPRC Score}  \\ \midrule
			\multirow{3}{*}{Score Methods}	& SAPS-II  	&& 0.8005	& 0.0080  	& 0.3625	& 0.0065 	&& 0.8030	& 0.0132  	& 0.3448	& 0.0219  \\
			& New SAPS-II 	&& 0.8217	& 0.0047  	& 0.4037	& 0.0069 	&& 0.8226	& 0.0129  	& 0.3873	& 0.0163  \\
			& SOFA  	&& 0.7263	& 0.0100  	& 0.3273	& 0.0067 	&& 0.7309	& 0.0105  	& 0.2996	& 0.0199  \\ \midrule
			\multirow{12}{*}{Super Learner}	& SL.glm 	&& 0.8212	& 0.0052  	& 0.4018	& 0.0068 	&& 0.8227	& 0.0132  	& 0.3883	& 0.0170  \\
			& SL.gbm  	&& 0.8405	& 0.0056  	& 0.4377	& 0.0112 	&& 0.8414	& 0.0111  	& 0.4187	& 0.0315  \\
			& SL.nnet 	&& 0.8332	& 0.0041  	& 0.4182	& 0.0059 	&& 0.8309	& 0.0061  	& 0.3905	& 0.0253  \\
			& SL.ipredbagg  	&& 0.7567	& 0.0040  	& 0.3063	& 0.0098 	&& 0.7483	& 0.0167  	& 0.2921	& 0.0211  \\
			& SL.randomforest 	&& 0.7553	& 0.0058  	& 0.3005	& 0.0121 	&& 0.7538	& 0.0150  	& 0.2914	& 0.0154  \\
			& SuperLearner-I  	&& 0.8417	& 0.0052  	& 0.4387	& 0.0122 	&& 0.8415	& 0.0096  	& 0.4169	& 0.0305  \\ \cmidrule{2-12}
			& SL.glm  	&& 0.8027	& 0.0038  	& 0.3931	& 0.0105 	&& 0.8009	& 0.0149  	& 0.3683	& 0.0209  \\
			& SL.gbm  	&& 0.8581	& 0.0062  	& 0.4810	& 0.0126 	&& 0.8457	& 0.0080  	& 0.4349	& 0.0233  \\
			& SL.nnet 	&& 0.8461	& 0.0103  	& 0.4674	& 0.0173 	&& 0.8238	& 0.0157  	& 0.4017	& 0.0412  \\
			& SL.ipredbagg  	&& 0.7921	& 0.0077  	& 0.3850	& 0.0180 	&& 0.7782	& 0.0052  	& 0.3434	& 0.0213  \\
			& SL.randomforest 	&& 0.7930	& 0.0063  	& 0.3830	& 0.0091 	&& 0.7733	& 0.0115  	& 0.3455	& 0.0253  \\
			& SuperLearner-II 	&& \bfseries 0.8651	& \bfseries 0.0075 	& \bfseries 0.4964	& \bfseries 0.0135  	&& 0.8520	& 0.0101  	& 0.4493	& 0.0246  \\ \midrule
			\multirow{2}{*}{Deep Learning}	& FFN  	&& 0.8488	& 0.0082  	& 0.4702	& 0.0168 	&& 0.8326	& 0.0112  	& 0.4109	& 0.0193  \\
			& RNN  	&& 0.8456	& 0.0032	& 0.4505	& 0.0091	&& 0.8485	& 0.0090	& 0.4246	& 0.0214  \\
			& MMDL 	&& 0.8561	& 0.0045  	& 0.4764	& 0.0144 	&& \bfseries 0.8564	& \bfseries 0.0107 	& \bfseries 0.4520	& \bfseries 0.0305   \\ \bottomrule
		\end{tabular}
	}
\end{table}

Tables \ref{table:mimic3-mor-b}, \ref{table:mimic3-mor-c}, \ref{table:mimic2-mor-b}, and \ref{table:mimic2-mor-c} show the in-hospital mortality prediction task results on Features set B and C of MIMIC-III and MIMIC-III (CareVue) datasets on the 24-hour and 48-hour data.
We observe that: (i) Super Learner performs better than all algorithms used in SuperLearner library, (ii) On both the Feature Set B and Feature Set C, the deep learning model (MMDL) obtains the best results in terms of AUROC and AUPRC score, (iii) We can observe that the results on first 48-hour data are similar with those on first 24-hour data, showing that a longer record length helps little on the in-hospital mortality prediction task.

\begin{figure}[b]
	\centering
    \subfigure{
        \includegraphics[width=0.22\textwidth]{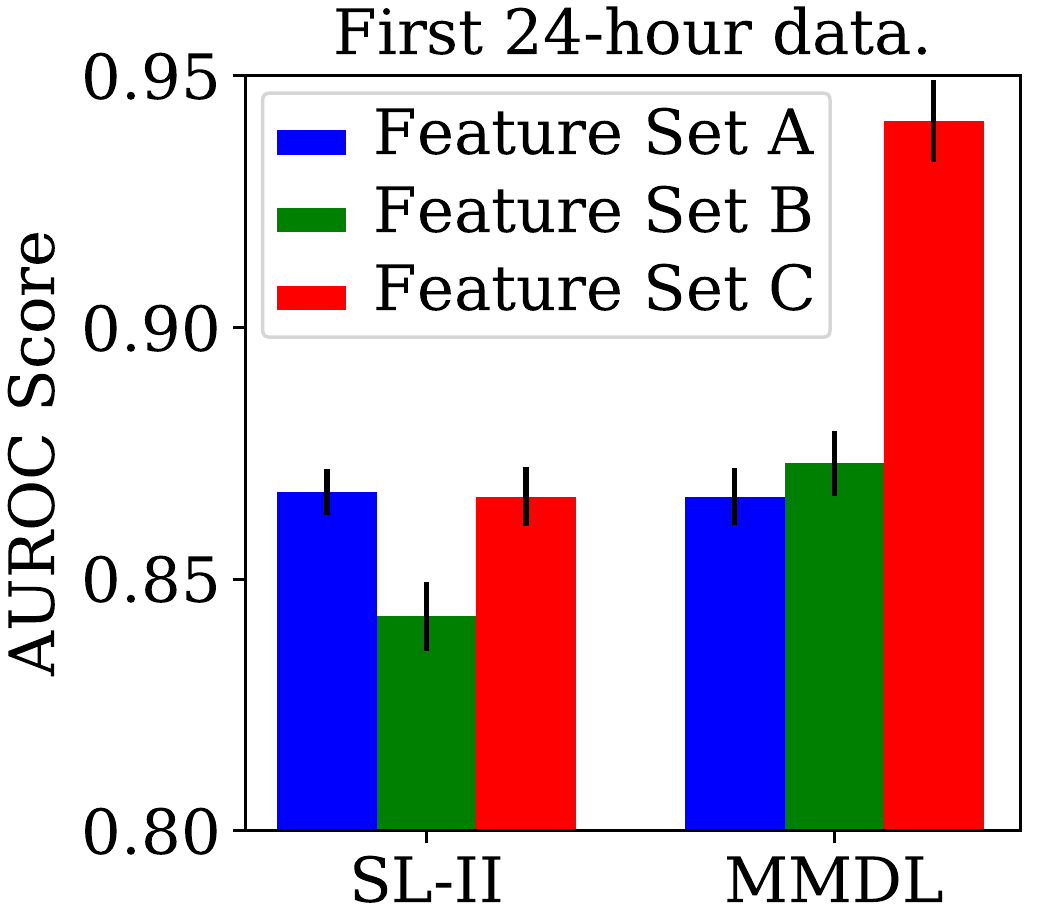}
    }
    \subfigure{
        \includegraphics[width=0.22\textwidth]{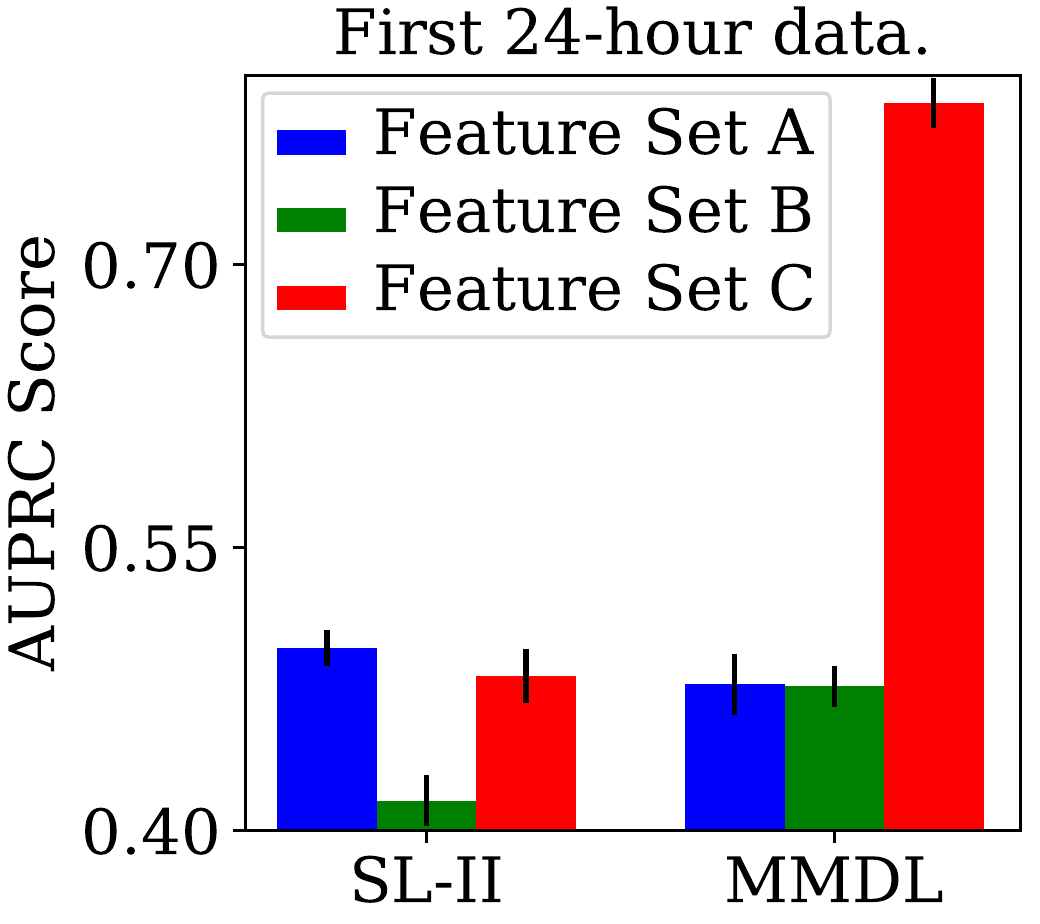}
    }
    \subfigure{
        \includegraphics[width=0.22\textwidth]{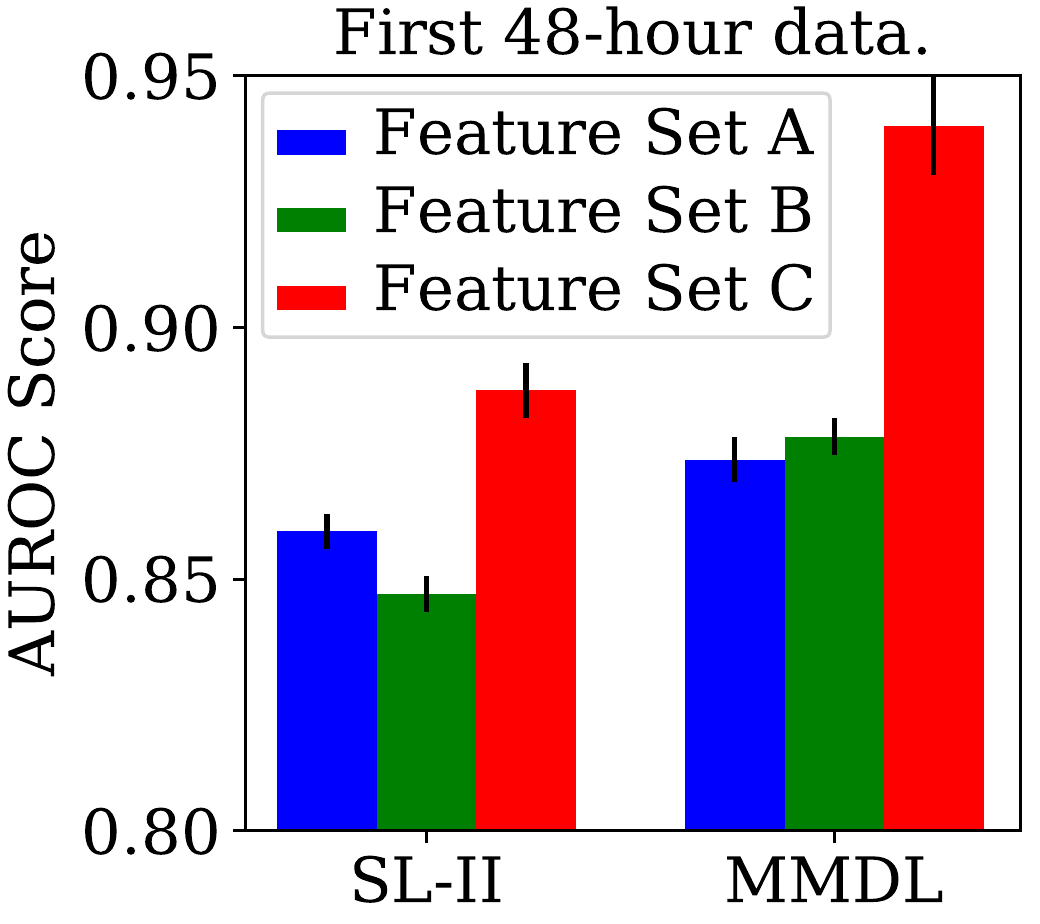}
    }
    \subfigure{
        \includegraphics[width=0.22\textwidth]{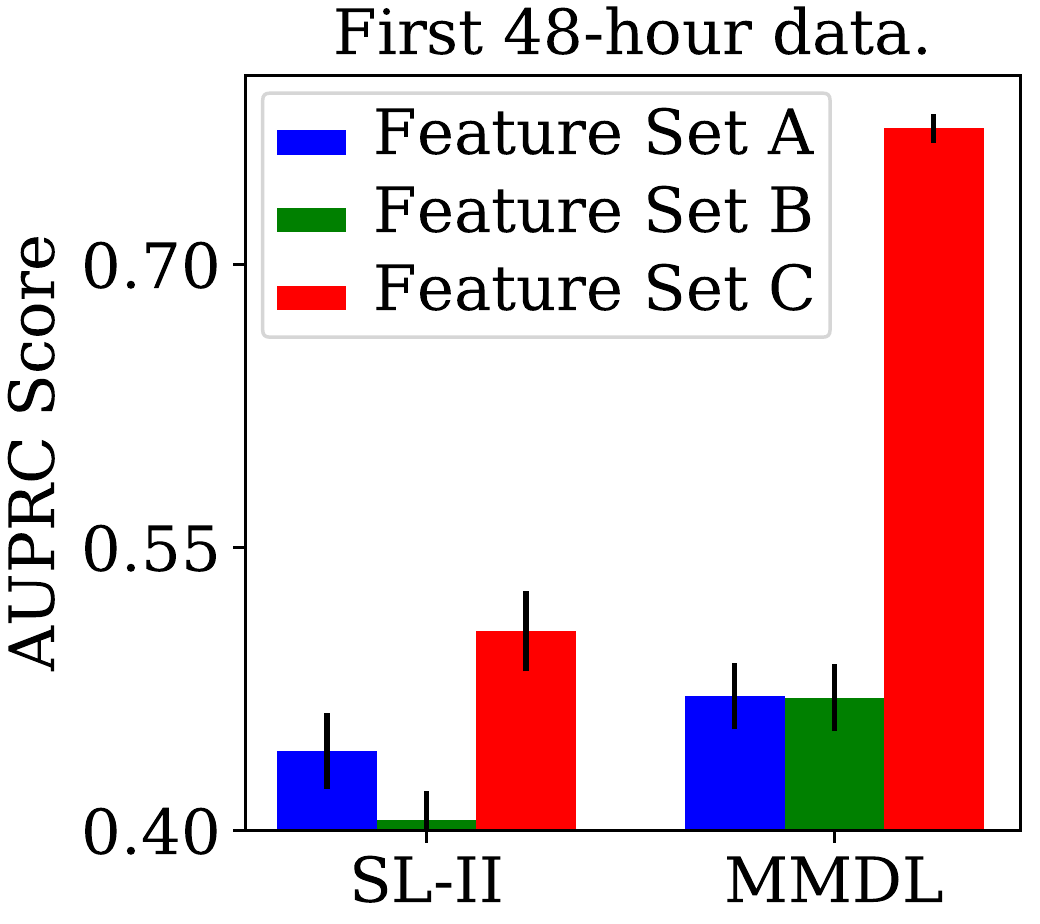}
    }
	\caption{In-hospital mortality task on MIMIC-III data.}
	\label{fig:mimic3-mor}
\end{figure}

\begin{figure}[b]
	\centering
    \subfigure{
        \includegraphics[width=0.22\textwidth]{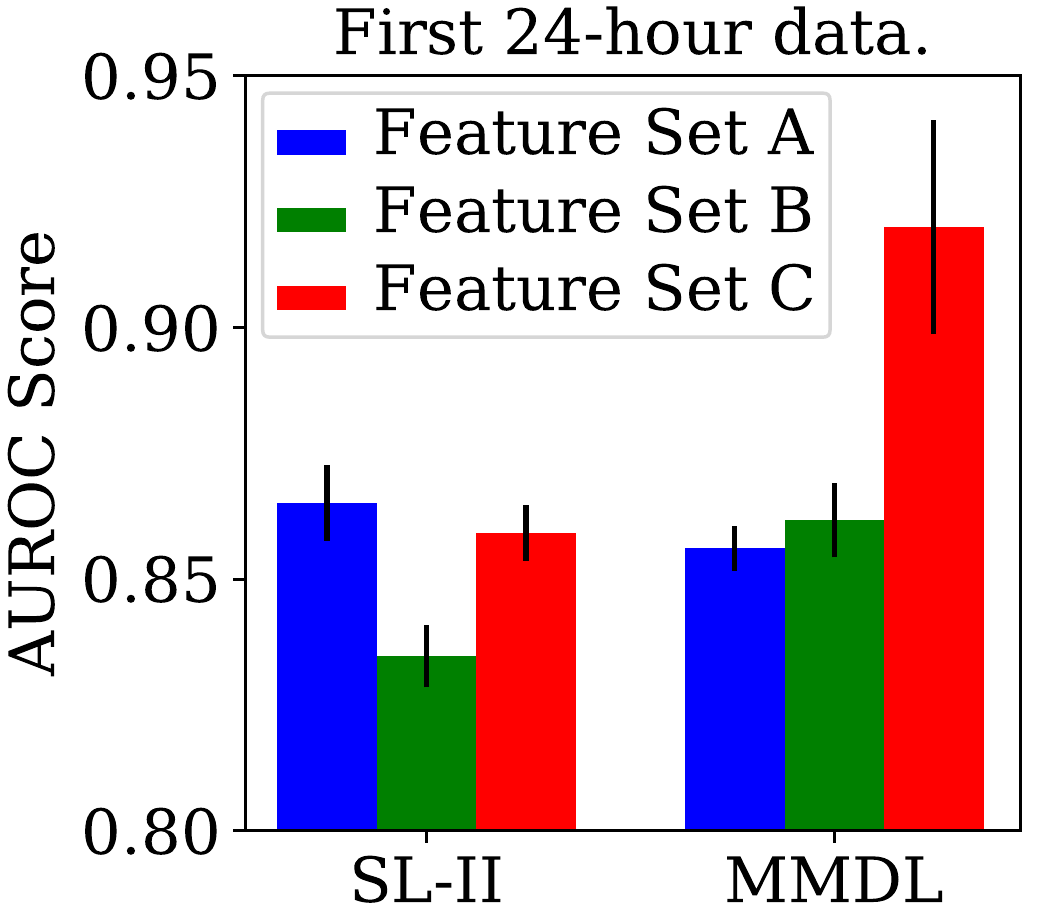}
    }
    \subfigure{
        \includegraphics[width=0.22\textwidth]{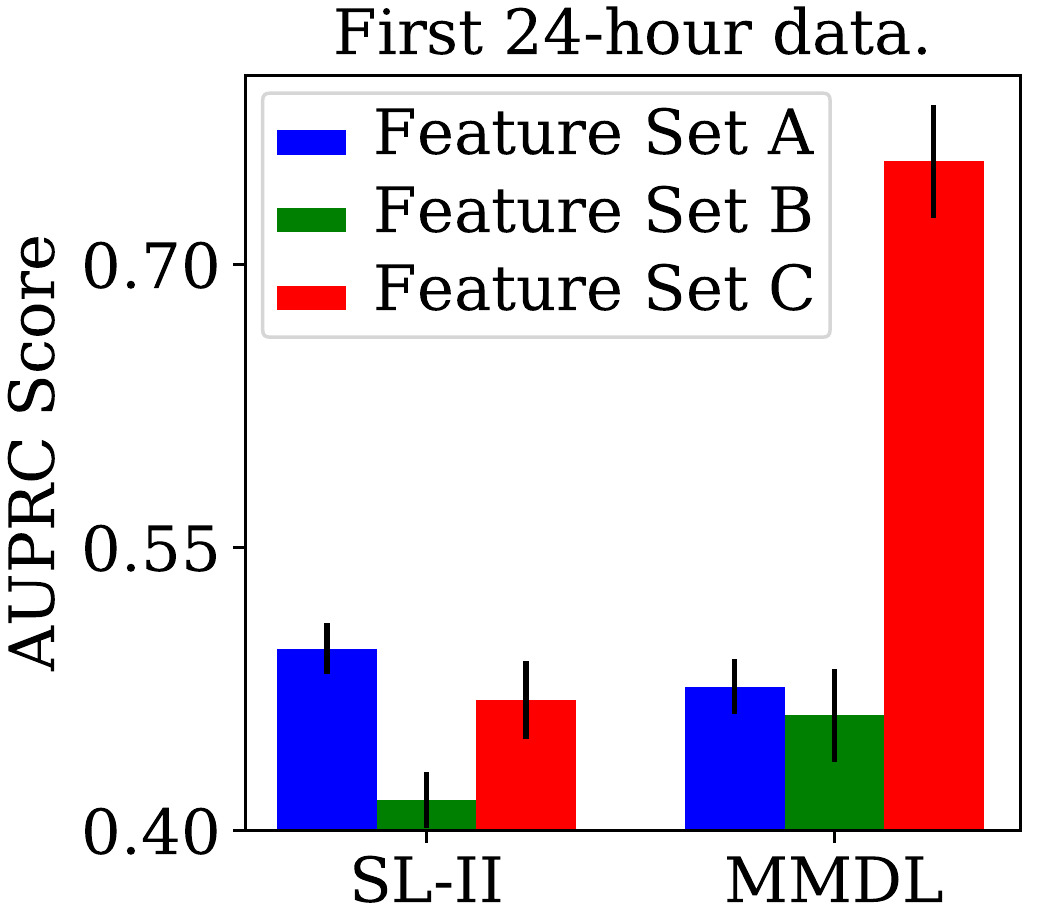}
    }
    \subfigure{
        \includegraphics[width=0.22\textwidth]{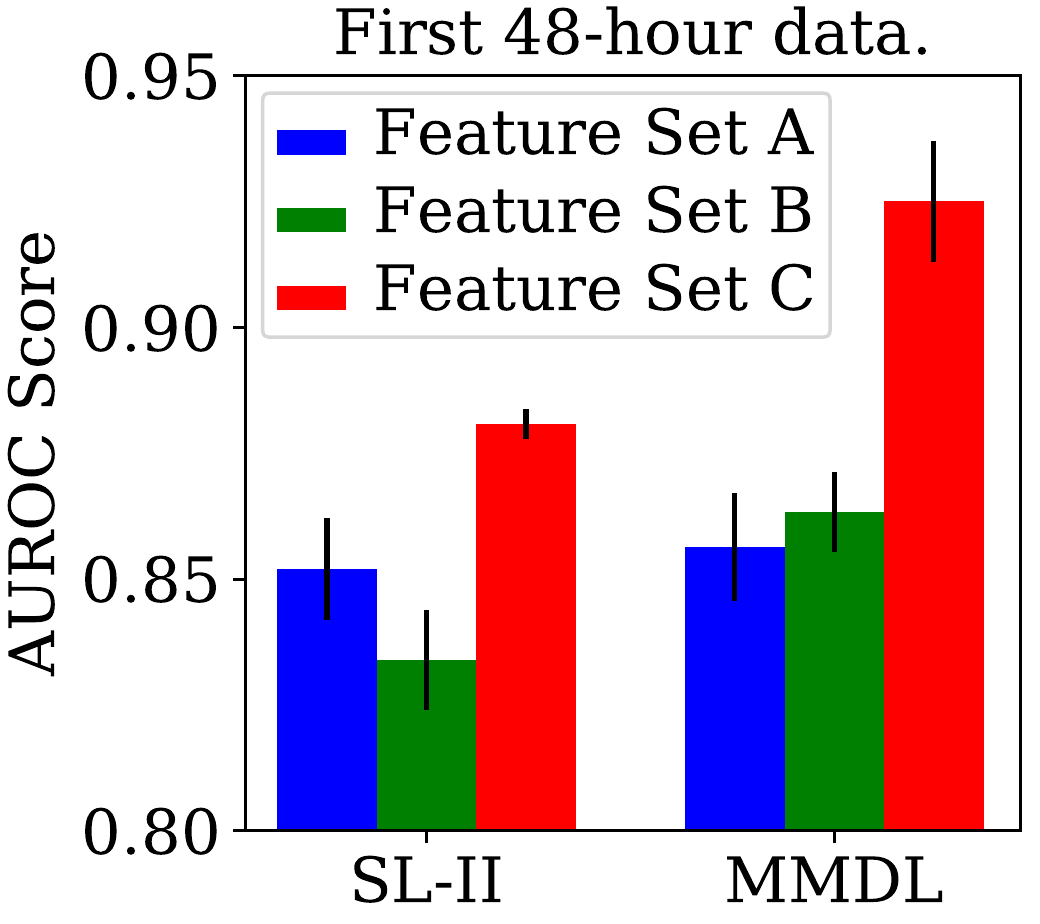}
    }
    \subfigure{
        \includegraphics[width=0.22\textwidth]{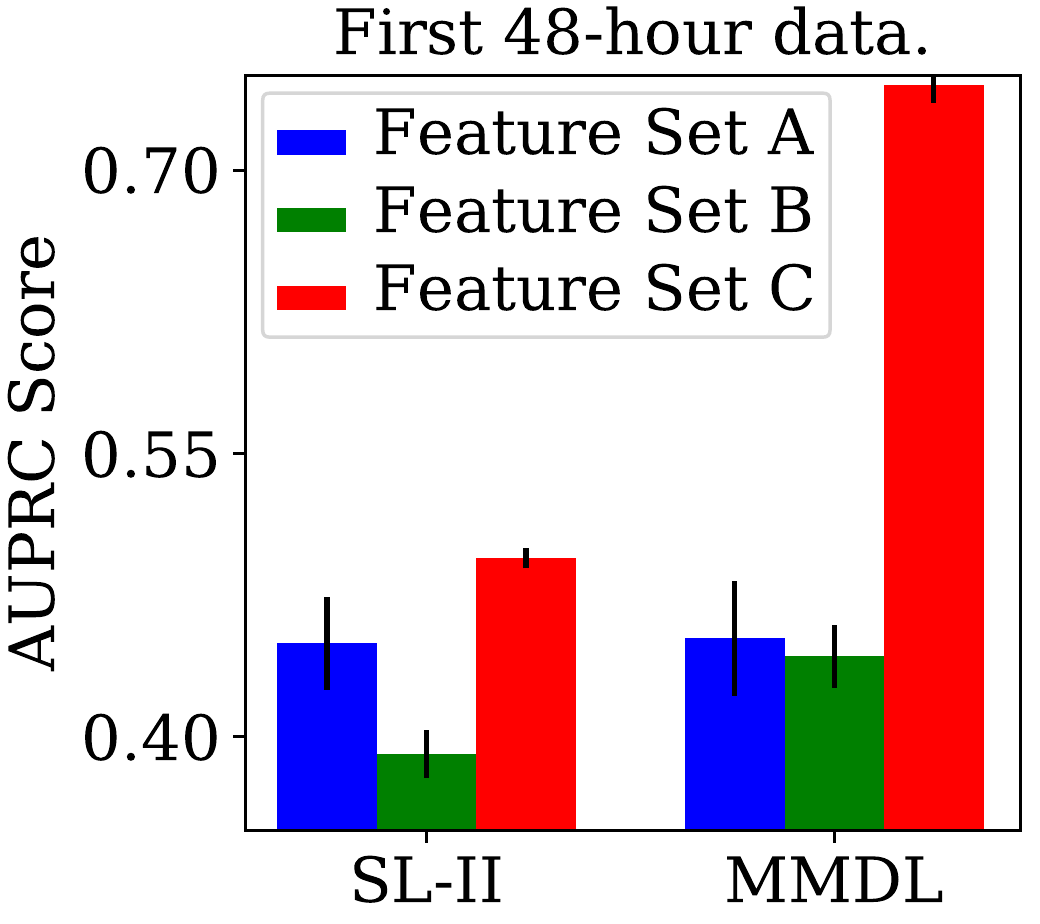}
    }
	\caption{In-hospital mortality task on MIMIC-III (CareVue) data.}
	\label{fig:mimic2-mor}
\end{figure}

From the in-hospital mortality prediction task results, we make the following observations: (i) Deep learning models (MMDL) outperform all the other models when the raw features (Feature set B and C) are used for evaluation, (ii) All the models perform much better when more features are used for prediction, i.e. models perform better on Feature set C which has 135 raw features compared to Feature Set B which has 20 features. This implies that deep models can learn better feature representations from multiple data modalities (instead of using hand-picked features as in Feature Set A) which results in obtaining better prediction results on the in-hospital mortality benchmark task.
The comparisons of SuperLearner-II and MMDL on three feature sets shown in Figures \ref{fig:mimic3-mor} and \ref{fig:mimic2-mor} validate our observations. From these figues, we see that on Feature set C, deep learning models obtain around 7-8\% and 50\% improvement over SuperLearner models for AUROC and AUPRC respectively. Also, deep learning models obtain 8\% improvement for Feature set C compared to Feature set A.

\begin{table}[t]
    \renewcommand{\arraystretch}{1.2}
	\centering
	\caption{In-hospital mortality task on MIMIC-III using feature set B.}
	\label{table:mimic3-mor-b}
	\resizebox{\textwidth}{!}{
		\begin{tabular}{@{}ll *2{l*2{S[table-format=1.4]@{$\ \pm\ $}S[table-format=1.4]}} @{}}
			\toprule
			\multirow{2}{*}{\textbf{Method}} 	&  \multirow{2}{*}{\textbf{Algorithm}} 	&&  \multicolumn{4}{c}{\textbf{Feature Set B, 24-hour data}} 	&&  \multicolumn{4}{c}{\textbf{Feature Set B, 48-hour data}} \\ \cmidrule{4-7} \cmidrule{9-12}
			& &	&  \multicolumn{2}{c}{AUROC Score} 	&  \multicolumn{2}{c}{AUPRC Score} 	&&  \multicolumn{2}{c}{AUROC Score}	&  \multicolumn{2}{c}{AUPRC Score}  \\ \midrule
			\multirow{6}{*}{Super Learner}  & SL.glm                  && 0.7745 & 0.0055          & 0.3134 & 0.0112          && 0.7869 & 0.0015          & 0.3103 & 0.0212          \\
			& SL.gbm                  && 0.8381 & 0.0057          & 0.4059 & 0.0156          && 0.8398 & 0.0044          & 0.3932 & 0.0155          \\
			& SL.nnet                 && 0.8170 & 0.0036          & 0.3650 & 0.0124          && 0.8232 & 0.0074          & 0.3591 & 0.0135          \\
			& SL.ipredbagg            && 0.7641 & 0.0070          & 0.3127 & 0.0085          && 0.7627 & 0.0118          & 0.3011 & 0.0140          \\
			& SL.randomforest         && 0.7582 & 0.0080          & 0.3100 & 0.0116          && 0.7604 & 0.0042          & 0.2895 & 0.0138          \\
			& SuperLearner-II         && 0.8426 & 0.0068          & 0.4160 & 0.0136          && 0.8471 & 0.0036          & 0.4055 & 0.0155          \\ \midrule
			\multirow{1}{*}{Deep Learning}
			& MMDL                && \bfseries 0.8730 & \bfseries 0.0065 & \bfseries 0.4765 & \bfseries 0.0109 && \bfseries 0.8783 & \bfseries 0.0037 & \bfseries 0.4706 & \bfseries 0.0178 \\ \bottomrule
		\end{tabular}
	}
\end{table}

\begin{table}[t]
    \renewcommand{\arraystretch}{1.2}
	\centering
	\caption{In-hospital mortality task on MIMIC-III using feature set C.}
	\label{table:mimic3-mor-c}
	\resizebox{\textwidth}{!}{
		\begin{tabular}{@{}ll *2{l*2{S[table-format=1.4]@{$\ \pm\ $}S[table-format=1.4]}} @{}}
			\toprule
			\multirow{2}{*}{\textbf{Method}} 	&  \multirow{2}{*}{\textbf{Algorithm}} 	&&  \multicolumn{4}{c}{\textbf{Feature Set C, 24-hour data}} 	&&  \multicolumn{4}{c}{\textbf{Feature Set C, 48-hour data}} \\ \cmidrule{4-7} \cmidrule{9-12}
			& &	&  \multicolumn{2}{c}{AUROC Score} 	&  \multicolumn{2}{c}{AUPRC Score} 	&&  \multicolumn{2}{c}{AUROC Score}	&  \multicolumn{2}{c}{AUPRC Score}  \\ \midrule
			\multirow{6}{*}{Super Learner}  & SL.glm                  && 0.8341 & 0.0072          & 0.4045 & 0.0164          && 0.8594 & 0.0079          & 0.4254 & 0.0196          \\
			& SL.gbm                  && 0.8628 & 0.0056          & 0.4705 & 0.0138          && 0.8833 & 0.0054          & 0.4954 & 0.0223          \\
			& SL.nnet                 && 0.7568 & 0.0106          & 0.3424 & 0.0139          && 0.7973 & 0.0060          & 0.3690 & 0.0154          \\
			& SL.ipredbagg            && 0.7895 & 0.0077          & 0.3664 & 0.0099          && 0.8074 & 0.0100          & 0.3796 & 0.0282          \\
			& SL.randomforest         && 0.7720 & 0.0054          & 0.3427 & 0.0045          && 0.7945 & 0.0081          & 0.3616 & 0.0143          \\
			& SuperLearner-II   	&& 0.8664 & 0.0058          & 0.4821 & 0.0142          && 0.8875 & 0.0055          & 0.5059 & 0.0214          \\ \midrule
			\multirow{1}{*}{Deep Learning}
			& MMDL                	&& \bfseries 0.9410 & \bfseries 0.0082 & \bfseries 0.7857 & \bfseries 0.0132 && \bfseries 0.9401 & \bfseries 0.0099 & \bfseries 0.7721 & \bfseries 0.0078 \\ \bottomrule
	\end{tabular}}
\end{table}

\begin{table}[t]
    \renewcommand{\arraystretch}{1.2}
	\centering
	\caption{In-hospital mortality task on MIMIC-III (CareVue) using feature set B.}
	\label{table:mimic2-mor-b}
	\resizebox{\textwidth}{!}{
		\begin{tabular}{@{}ll *2{l*2{S[table-format=1.4]@{$\ \pm\ $}S[table-format=1.4]}} @{}}
			\toprule
			\multirow{2}{*}{\textbf{Method}} 	&  \multirow{2}{*}{\textbf{Algorithm}} 	&&  \multicolumn{4}{c}{\textbf{Feature Set B, 24-hour data}} 	&&  \multicolumn{4}{c}{\textbf{Feature Set B, 48-hour data}} \\ \cmidrule{4-7} \cmidrule{9-12}
			& &	&  \multicolumn{2}{c}{AUROC Score} 	&  \multicolumn{2}{c}{AUPRC Score} 	&&  \multicolumn{2}{c}{AUROC Score}	&  \multicolumn{2}{c}{AUPRC Score}  \\ \midrule
			\multirow{6}{*}{Super Learner}  & SL.glm                  && 0.7724 & 0.0129          & 0.3177 & 0.0185          && 0.7872 & 0.0139          & 0.3253 & 0.0236          \\
			& SL.gbm                  && 0.8284 & 0.0064          & 0.4048 & 0.0110          && 0.8298 & 0.0089          & 0.3823 & 0.0189          \\
			& SL.nnet                 && 0.8133 & 0.0060          & 0.3740 & 0.0169          && 0.7976 & 0.0201          & 0.3339 & 0.0305         \\
			& SL.ipredbagg            && 0.7525 & 0.0066          & 0.3158 & 0.0152          && 0.7536 & 0.0136          & 0.2991 & 0.0197          \\
			& SL.randomforest         && 0.7497 & 0.0118          & 0.3123 & 0.0110          && 0.7502 & 0.0094          & 0.2952 & 0.0074          \\
			& SuperLearner-II 		&& 0.8347 & 0.0062          & 0.4164 & 0.0148          && 0.8340 & 0.0099          & 0.3907 & 0.0128          \\ \midrule
			\multirow{1}{*}{Deep Learning}
			& MMDL                && \bfseries 0.8617 & \bfseries 0.0074 & \bfseries 0.4612 & \bfseries 0.0245 && \bfseries 0.8633 & \bfseries 0.0080 & \bfseries 0.4425 & \bfseries 0.0166 \\ \bottomrule
	\end{tabular}}
\end{table}

\begin{table}[t]
    \renewcommand{\arraystretch}{1.2}
	\centering
	\caption{In-hospital mortality task on MIMIC-III (CareVue) using feature set C.}
	\label{table:mimic2-mor-c}
	\resizebox{\textwidth}{!}{
		\begin{tabular}{@{}ll *2{l*2{S[table-format=1.4]@{$\ \pm\ $}S[table-format=1.4]}} @{}}
			\toprule
			\multirow{2}{*}{\textbf{Method}} 	&  \multirow{2}{*}{\textbf{Algorithm}} 	&&  \multicolumn{4}{c}{\textbf{Feature Set C, 24-hour data}} 	&&  \multicolumn{4}{c}{\textbf{Feature Set C, 48-hour data}} \\ \cmidrule{4-7} \cmidrule{9-12}
			& &	&  \multicolumn{2}{c}{AUROC Score} 	&  \multicolumn{2}{c}{AUPRC Score} 	&&  \multicolumn{2}{c}{AUROC Score}	&  \multicolumn{2}{c}{AUPRC Score}  \\ \midrule
			\multirow{6}{*}{Super Learner}  & SL.glm                  && 0.8282 & 0.0089          & 0.3931 & 0.0136          && 0.8497 & 0.0088          & 0.4128 & 0.0087          \\
			& SL.gbm                  && 0.8550 & 0.0049          & 0.4605 & 0.0209          && 0.8753 & 0.0026          & 0.4846 & 0.0042          \\
			& SL.nnet                 && 0.7412 & 0.0067          & 0.3216 & 0.0127          && 0.7946 & 0.0132          & 0.3535 & 0.0198          \\
			& SL.ipredbagg            && 0.7737 & 0.0064          & 0.3216 & 0.0127          && 0.7976 & 0.0065          & 0.3696 & 0.0127          \\
			& SL.randomforest         && 0.7690 & 0.0080          & 0.3435 & 0.0184          && 0.7900 & 0.0101          & 0.3558 & 0.0132          \\
			& SuperLearner-II 		&& 0.8592 & 0.0055          & 0.4694 & 0.0207          && 0.8808 & 0.0030          & 0.4945 & 0.0054          \\ \midrule
			\multirow{1}{*}{Deep Learning}
			& MMDL                && \bfseries 0.9200 & \bfseries 0.0213 & \bfseries 0.7546 & \bfseries 0.0297 && \bfseries 0.9251 & \bfseries 0.0120 & \bfseries 0.7451 & \bfseries 0.0093 \\ \bottomrule
	\end{tabular}}
\end{table}

\paragraph{Short-term and Long-term Mortality Prediction}
Tables \ref{table:mimic3-24h-mor-auroc}, \ref{table:mimic3-24h-mor-auprc}, \ref{table:mimic3-48h-mor-auroc}, and \ref{table:mimic3-48h-mor-auprc} 
show the short-term and long-term mortality prediction task results on all the feature sets of MIMIC-III dataset on the 24-hour and 48-hour data.
We observe that: (i) Super Learner-II and MMDL deep learning models have similar performance on the feature set A for both short-term and long-term mortality prediction, and both these algorithms perform better than all other prediction algorithms, (ii) On both the Feature Set B and Feature Set C, the MMDL deep learning model consistently obtains the best results in terms of AUROC and AUPRC score, (iii) all models obtain better AUPRC scores on the long-term mortality prediction task compared to the short-term mortality prediction task.

\begin{table}[t]
 \renewcommand{\arraystretch}{1.2}
	\centering
	\caption{AUROC scores of short-term and long-term mortality prediction tasks on MIMIC-III with 24-hour data.}
	\label{table:mimic3-24h-mor-auroc}
	\resizebox{\textwidth}{!}{
	\begin{tabular}{@{}ll*4{lS[table-format=1.4]@{$\ \pm\ $}S[table-format=1.4]}@{}}
		\toprule
		\multirow{2}{*}{\textbf{Feature Set}}	& \multirow{2}{*}{\textbf{Algorithm}}	&& \multicolumn{11}{c}{\textbf{AUROC Score}} \\ \cmidrule{4-14}
		&&& \multicolumn{2}{c}{2-day Mortality}	&& \multicolumn{2}{c}{3-day Mortality}	&& \multicolumn{2}{c}{30-day Mortality}	&& \multicolumn{2}{c}{1-year Mortality} \\ \midrule
		\multirow{8}{*}{Feature Set A}	& SAPS-II Score	
							&& 0.8453	& 0.0088	&& 0.8218	& 0.0057	&& 0.7921	& 0.0051	&& 0.7614	& 0.0035 \\
		& New SAPS-II Score	&& 0.8575	& 0.0075	&& 0.8370	& 0.0053	&& 0.8148	& 0.0035	&& 0.8042	& 0.0013 \\
		& SOFA Score		&& 0.7559	& 0.0276	&& 0.7412	& 0.0076	&& 0.7041	& 0.0074	&& 0.6611	& 0.0036 \\
		& SuperLearner-I	&& 0.8808	& 0.0063	&& 0.8627	& 0.0079	&& 0.8384	& 0.0031	&& 0.8260	& 0.0019 \\
		& SuperLearner-II	&& 0.8851	& 0.0105	&& 0.8770	& 0.0094	&& 0.8620	& 0.0063	&& 0.8467	& 0.0022 \\
		& FFN				&& 0.8673	& 0.0069	&& 0.8493	& 0.0128	&& 0.8475	& 0.0050	&& 0.8390	& 0.0019 \\
		& RNN				&& 0.8773	& 0.0117	&& 0.8612	& 0.0083	&& 0.8326	& 0.0085	&& 0.7958	& 0.0026 \\
		& MMDL				&& 0.8815	& 0.0102	&& 0.8725	& 0.0063	&& 0.8585	& 0.0059	&& 0.8450	& 0.0019 \\ \midrule
		\multirow{2}{*}{Feature Set B}	& SuperLearner-II	
							&& 0.8667	& 0.0097	&& 0.8535	& 0.0128	&& 0.8395	& 0.0031	&& 0.8347	& 0.0046 \\
		& MMDL				&& 0.8862	& 0.0059	&& 0.8769	& 0.0107	&& 0.8620	& 0.0072	&& 0.8452	& 0.0008 \\ \midrule
		\multirow{2}{*}{Feature Set C}	& SuperLearner-II	
							&& 0.8837	& 0.0047	&& 0.8746	& 0.0073	&& 0.8629	& 0.0033	&& 0.8589	& 0.0032 \\
		& MMDL				&& \bfseries 0.9084	& \bfseries 0.0207	&& \bfseries 0.9295	& \bfseries 0.0225	&& \bfseries 0.9169	& \bfseries 0.0054	&& \bfseries 0.8872	& \bfseries 0.0084 \\ \bottomrule
	\end{tabular}}
\end{table}

\begin{table}[t]
 \renewcommand{\arraystretch}{1.2}
	\centering
	\caption{AUPRC scores of short-term and long-term mortality prediction tasks on MIMIC-III with 24-hour data.}
	\label{table:mimic3-24h-mor-auprc}
	\resizebox{\textwidth}{!}{
	\begin{tabular}{@{}ll*4{lS[table-format=1.4]@{$\ \pm\ $}S[table-format=1.4]}@{}}
		\toprule
		\multirow{2}{*}{\textbf{Feature Set}}	& \multirow{2}{*}{\textbf{Algorithm}}	&& \multicolumn{11}{c}{\textbf{AUPRC Score}} \\ \cmidrule{4-14}
		&&& \multicolumn{2}{c}{2-day Mortality}	&& \multicolumn{2}{c}{3-day Mortality}	&& \multicolumn{2}{c}{30-day Mortality}	&& \multicolumn{2}{c}{1-year Mortality} \\ \midrule
		\multirow{8}{*}{Feature Set A}	& SAPS-II Score	
							&& 0.1361	& 0.0153	&& 0.1730	& 0.0214	&& 0.4140	& 0.0131	&& 0.5084	& 0.0154 \\
		& New SAPS-II Score	&& 0.1587	& 0.0226	&& 0.1919	& 0.0234	&& 0.4589	& 0.0125	&& 0.5778	& 0.0109 \\
		& SOFA Score		&& 0.1027	& 0.0278	&& 0.1373	& 0.0201	&& 0.3497	& 0.0167	&& 0.4176	& 0.0088 \\
		& SuperLearner-I	&& 0.1967	& 0.0205	&& 0.2219	& 0.0263	&& 0.5053	& 0.0173	&& 0.6258	& 0.0073 \\
		& SuperLearner-II	&& 0.2463	& 0.0111	&& 0.2775	& 0.0382	&& 0.5652	& 0.0186	&& 0.6609	& 0.0090 \\
		& FFN				&& 0.2429	& 0.0332	&& 0.2449	& 0.0315	&& 0.5367	& 0.0199	&& 0.6453	& 0.0081 \\
		& RNN				&& 0.2491	& 0.0293	&& 0.2752	& 0.0164	&& 0.5028	& 0.0178	&& 0.5725	& 0.0062 \\
		& MMDL				&& 0.2529	& 0.0338	&& 0.2839	& 0.0207	&& 0.5483	& 0.0187	&& 0.6485	& 0.0099 \\ \midrule
		\multirow{2}{*}{Feature Set B}	& SuperLearner-II	
							&& 0.1767	& 0.0319	&& 0.2173	& 0.0266	&& 0.4926	& 0.0090	&& 0.6328	& 0.0100 \\
		& MMDL				&& 0.2475	& 0.0364	&& 0.1863	& 0.0273	&& 0.5458	& 0.0231	&& 0.6457	& 0.0082 \\ \midrule
		\multirow{2}{*}{Feature Set C}	& SuperLearner-II	
							&& 0.2048	& 0.0085	&& 0.2717	& 0.0321	&& 0.5530	& 0.0096	&& 0.6764	& 0.0056 \\
		& MMDL				&& \bfseries 0.3831	& \bfseries 0.0336	&& \bfseries 0.5139	& \bfseries 0.0193	&& \bfseries 0.7668	& \bfseries 0.0170	&& \bfseries 0.7690	& \bfseries 0.0077 \\ \bottomrule
	\end{tabular}}
\end{table}

\begin{table}[t]
 \renewcommand{\arraystretch}{1.2}
	\centering
	\caption{AUROC scores of short-term and long-term mortality prediction tasks on MIMIC-III with 48-hour data.}
	\label{table:mimic3-48h-mor-auroc}
	\resizebox{\textwidth}{!}{
	\begin{tabular}{@{}ll*3{lS[table-format=1.4]@{$\ \pm\ $}S[table-format=1.4]}@{}}
		\toprule
		\multirow{2}{*}{\textbf{Feature Set}}	& \multirow{2}{*}{\textbf{Algorithm}}	&& \multicolumn{8}{c}{\textbf{AUROC Score}} \\ \cmidrule{4-11}
		&&& \multicolumn{2}{c}{3-day Mortality}	&& \multicolumn{2}{c}{30-day Mortality}	&& \multicolumn{2}{c}{1-year Mortality} \\ \midrule
			\multirow{8}{*}{Feature Set A}	& SAPS-II Score	
								&& 0.8366	& 0.0109	&& 0.7841	& 0.0072	&& 0.7490	& 0.0041 \\
			& New SAPS-II Score	&& 0.8471	& 0.0072	&& 0.8104	& 0.0047	&& 0.7991	& 0.0037 \\
			& SOFA Score		&& 0.7465	& 0.0179	&& 0.6953	& 0.0104	&& 0.6454	& 0.0052 \\
			& SuperLearner-I	&& 0.8675	& 0.0046	&& 0.8364	& 0.0033	&& 0.8222	& 0.0047 \\
			& SuperLearner-II	&& 0.8706	& 0.0095	&& 0.8531	& 0.0043	&& 0.8409	& 0.0031 \\
			& FFN				&& 0.8466	& 0.0186	&& 0.8385	& 0.0061	&& 0.8309	& 0.0048 \\
			& RNN				&& 0.8633	& 0.0116	&& 0.8374	& 0.0087	&& 0.7966	& 0.0036 \\
			& MMDL				&& 0.8596	& 0.0124	&& 0.8612	& 0.0059	&& 0.8418	& 0.0049 \\ \midrule
			\multirow{2}{*}{Feature Set B}	& SuperLearner-II	
								&& 0.8448	& 0.0162	&& 0.8427	& 0.0071	&& 0.8360	& 0.0057 \\
			& MMDL				&& 0.8682	& 0.0240	&& 0.8628	& 0.0111	&& 0.8438	& 0.0053 \\ \midrule
			\multirow{2}{*}{Feature Set C}	& SuperLearner-II
								&& 0.8473	& 0.0114	&& 0.8802	& 0.0037	&& 0.8673	& 0.0051 \\
			& MMDL				&& \bfseries 0.8713	& \bfseries 0.0494	&& \bfseries 0.9173	& \bfseries 0.0064	&& \bfseries 0.8702	& \bfseries 0.0054 \\ \bottomrule
		\end{tabular}
	}
\end{table}

\begin{table}[t]
 \renewcommand{\arraystretch}{1.2}
	\centering
	\caption{AUPRC scores of short-term and long-term mortality prediction tasks on MIMIC-III with 48-hour data.}
	\label{table:mimic3-48h-mor-auprc}
	\resizebox{\textwidth}{!}{
	\begin{tabular}{@{}ll*3{lS[table-format=1.4]@{$\ \pm\ $}S[table-format=1.4]}@{}}
		\toprule
		\multirow{2}{*}{\textbf{Feature Set}}	& \multirow{2}{*}{\textbf{Algorithm}}	&& \multicolumn{8}{c}{\textbf{AUPRC Score}} \\ \cmidrule{4-11}
		&&& \multicolumn{2}{c}{3-day Mortality}	&& \multicolumn{2}{c}{30-day Mortality}	&& \multicolumn{2}{c}{1-year Mortality} \\ \midrule
			\multirow{8}{*}{Feature Set A}	& SAPS-II Score	
								&& 0.1082	& 0.0150	&& 0.3849	& 0.0118	&& 0.4845	& 0.0092 \\
			& New SAPS-II Score	&& 0.1307	& 0.0225	&& 0.4342	& 0.0119	&& 0.5647	& 0.0090 \\
			& SOFA Score		&& 0.0663	& 0.0092	&& 0.3156	& 0.0143	&& 0.3898	& 0.0115 \\
			& SuperLearner-I	&& 0.1344	& 0.0247	&& 0.4898	& 0.0139	&& 0.6171	& 0.0088 \\
			& SuperLearner-II	&& 0.1955	& 0.0245	&& 0.5255	& 0.0152	&& 0.6448	& 0.0084 \\
			& FFN				&& 0.1672	& 0.0331	&& 0.4962	& 0.0153	&& 0.6272	& 0.0116 \\
			& RNN				&& \bfseries 0.2371	& \bfseries 0.0336	&& 0.4974	& 0.0149	&& 0.5691	& 0.0080 \\
			& MMDL				&& 0.2131	& 0.0344	&& 0.5423	& 0.0164	&& 0.6421	& 0.0116 \\ \midrule
			\multirow{2}{*}{Feature Set B}	& SuperLearner-II	
								&& 0.1225	& 0.0286	&& 0.4892	& 0.0197	&& 0.6297	& 0.0067 \\
			& MMDL				&& 0.1659	& 0.0434	&& 0.5290	& 0.0372	&& 0.6444	& 0.0133 \\ \midrule
			\multirow{2}{*}{Feature Set C}	& SuperLearner-II	
								&& 0.0771	& 0.0125	&& 0.5479	& 0.0079	&& 0.6870	& 0.0038 \\
			& MMDL				&& 0.1510	& 0.0246	&& \bfseries 0.7314	& \bfseries 0.0149	&& \bfseries 0.7344	& \bfseries 0.0062 \\ \bottomrule
		\end{tabular}
	}
\end{table}

\subsubsection{ICD-9 Code Prediction Task Evaluation}
Tables \ref{tab:icd9-mimic-24h-AUPRC-scores} and \ref{tab:icd9-mimic-24h-AUROC-scores} show the performance (AUPRC and AUROC scores) of all methods for the first 24-hour data of MIMIC-III on ICD-9 code prediction task.
We observe that the MMDL deep models trained on Feature Set C outperforms Super Learner models trained on Feature Sets A, B, and C on almost all the ICD-9 Code prediction task, and on an average obtains 4-5\% improvement.

\begin{sidewaystable}
	\renewcommand{\arraystretch}{1.3}
	\centering
	\caption{ICD-9 code prediction AUPRC scores on MIMIC-III with first 24-hour data.}
	\label{tab:icd9-mimic-24h-AUPRC-scores}
	\resizebox{\textwidth}{!}{
		\begin{tabular}{p{0.06\textwidth}*7{lS[table-format=1.4]@{$\ \pm\ $}S[table-format=1.4]}}
			\toprule
			 \rowcell{3}{ICD-9 \\ Task}	&& \multicolumn{11}{c}{Super Learner}	&& \multicolumn{8}{c}{Deep Learning}	\\	
			\cmidrule{3-13} \cmidrule{15-22}
			&& \multicolumn{2}{c}{Super Learner-I}	&& \multicolumn{2}{c}{Super Learner-II}	&& \multicolumn{2}{c}{Super Learner-II}	&& \multicolumn{2}{c}{Super Learner-II}	&& \multicolumn{2}{c}{FFN}	&& \multicolumn{2}{c}{RNN}	&& \multicolumn{2}{c}{MMDL} \\	
			&& \multicolumn{2}{c}{on Feature Set A}	&& \multicolumn{2}{c}{on Feature Set A}	&& \multicolumn{2}{c}{on Feature Set B}	&& \multicolumn{2}{c}{on Feature Set C}	&& \multicolumn{2}{c}{on Feature Set C}	&& \multicolumn{2}{c}{on Feature Set C}	&& \multicolumn{2}{c}{on Feature Set C} \\ \midrule	
			 1	&& 0.5356	& 0.0027	&& 0.5861	& 0.0090	&& 0.5695	& 0.0075	&& 0.6273	& 0.0077	&& 0.5807	& 0.0072	&& 0.5978	& 0.0211	&& \bfseries 0.6491	& \bfseries 0.0121 \\	
			 2	&& 0.7290	& 0.0141	&& 0.7512	& 0.0130	&& 0.7478	& 0.0097	&& 0.7756	& 0.0134	&& 0.7422	& 0.0064	&& 0.4715	& 0.0132	&& \bfseries 0.8024	& \bfseries 0.0192 \\	
			 3	&& 0.8095	& 0.0057	&& 0.8235	& 0.0054	&& 0.8302	& 0.0057	&& 0.8631	& 0.0042	&& 0.8377	& 0.0055	&& 0.8614	& 0.0022	&& \bfseries 0.8690	& \bfseries 0.0115 \\	
			 4	&& 0.5454	& 0.0072	&& 0.5850	& 0.0065	&& 0.5836	& 0.0037	&& 0.6882	& 0.0128	&& 0.6556	& 0.0162	&& 0.6812	& 0.0125	&& \bfseries 0.7149	& \bfseries 0.0180 \\	
			 5	&& 0.4714	& 0.0047	&& 0.5168	& 0.0073	&& 0.5218	& 0.0044	&& \bfseries 0.5624	& \bfseries 0.0019	&& 0.5058	& 0.0100	&& 0.5220	& 0.0048	&& 0.5590	& 0.0212 \\	
			 6	&& 0.4112	& 0.0046	&& 0.4348	& 0.0064	&& 0.4507	& 0.0067	&& 0.5155	& 0.0074	&& 0.4566	& 0.0175	&& 0.5261	& 0.0076	&& \bfseries 0.5624	& \bfseries 0.0112 \\	
			 7	&& 0.9547	& 0.0026	&& 0.9596	& 0.0025	&& 0.9585	& 0.0019	&& 0.9701	& 0.0018	&& 0.9592	& 0.0013	&& 0.9552	& 0.0040	&& \bfseries 0.9729	& \bfseries 0.0021 \\	
			 8	&& 0.6960	& 0.0061	&& 0.7366	& 0.0045	&& 0.7306	& 0.0079	&& 0.7649	& 0.0038	&& 0.7493	& 0.0050	&& 0.8067	& 0.0062	&& \bfseries 0.8290	& \bfseries 0.0113 \\	
			 9	&& 0.5814	& 0.0080	&& 0.6350	& 0.0107	&& 0.6326	& 0.0087	&& 0.6961	& 0.0091	&& 0.6393	& 0.0033	&& 0.6748	& 0.0106	&& \bfseries 0.7034	& \bfseries 0.0148 \\	
			 10	&& 0.7213	& 0.0034	&& 0.7442	& 0.0047	&& 0.7328	& 0.0063	&& 0.7928	& 0.0053	&& 0.7715	& 0.0036	&& 0.8071	& 0.0020	&& \bfseries 0.8227	& \bfseries 0.0112 \\	
			 11	&& 0.0700	& 0.0180	&& 0.0965	& 0.0193	&& 0.0857	& 0.0082	&& 0.2303	& 0.0345	&& 0.1498	& 0.0565	&& 0.1623	& 0.0504	&& \bfseries 0.4888	& \bfseries 0.0631 \\	
			 12	&& 0.1629	& 0.0106	&& 0.1802	& 0.0109	&& 0.1862	& 0.0138	&& 0.2146	& 0.0213	&& 0.1930	& 0.0117	&& 0.2019	& 0.0149	&& \bfseries 0.3155	& \bfseries 0.0421 \\	
			 13	&& 0.2436	& 0.0070	&& 0.2551	& 0.0062	&& 0.2554	& 0.0051	&& 0.3144	& 0.0074	&& 0.2641	& 0.0078	&& 0.2945	& 0.0155	&& \bfseries 0.3435	& \bfseries 0.0293 \\	
			 14	&& 0.1318	& 0.0177	&& 0.1357	& 0.0219	&& 0.1370	& 0.0198	&& 0.1329	& 0.0183	&& 0.1134	& 0.0225	&& 0.0775	& 0.0068	&& \bfseries 0.1918	& \bfseries 0.0216 \\	
			 15	&& 0.4745	& 0.0061	&& 0.5036	& 0.0032	&& 0.4908	& 0.0061	&& 0.5343	& 0.0096	&& 0.4848	& 0.0096	&& 0.5182	& 0.0079	&& \bfseries 0.5564	& \bfseries 0.0195 \\	
			 16	&& 0.1164	& 0.0057	&& 0.1279	& 0.0082	&& 0.1264	& 0.0080	&& 0.1711	& 0.0097	&& 0.1488	& 0.0093	&& 0.1433	& 0.0104	&& \bfseries 0.2244	& \bfseries 0.0279 \\	
			 17	&& 0.0632	& 0.0113	&& 0.0649	& 0.0028	&& 0.0736	& 0.0048	&& 0.0913	& 0.0101	&& 0.0742	& 0.0055	&& 0.0604	& 0.0072	&& \bfseries 0.3700	& \bfseries 0.0577 \\	
			 18	&& 0.5934	& 0.0069	&& 0.6302	& 0.0053	&& 0.6300	& 0.0059	&& 0.6826	& 0.0072	&& 0.6430	& 0.0133	&& 0.6675	& 0.0068	&& \bfseries 0.7037	& \bfseries 0.0246 \\	
			 19	&& 0.5946	& 0.0029	&& 0.6205	& 0.0053	&& 0.6361	& 0.0065	&& 0.7138	& 0.0076	&& 0.6684	& 0.0041	&& 0.6951	& 0.0049	&& \bfseries 0.7184	& \bfseries 0.0086 \\	
			 20	&& 0.4820	& 0.0069	&& 0.5220	& 0.0128	&& 0.5382	& 0.0054	&& 0.6006	& 0.0103	&& 0.5506	& 0.0066	&& 0.5723	& 0.0041	&& \bfseries 0.6184	& \bfseries 0.0208 \\ \cmidrule{1-22}	
			 Average	&& 0.4694	& 0.0076	&& 0.4955	& 0.0083	&& 0.4959	& 0.0073	&& 0.5471	& 0.0102	&& 0.5094	& 0.0111	&& 0.5148	& 0.0107	&& \bfseries 0.6008	& \bfseries 0.0224 \\	
			\bottomrule
		\end{tabular}
	}
\end{sidewaystable}

\begin{sidewaystable}
	\renewcommand{\arraystretch}{1.3}
	\centering
	\caption{ICD-9 code prediction AUROC scores on MIMIC-III with first 24-hour data.}
	\label{tab:icd9-mimic-24h-AUROC-scores}
	\resizebox{\textwidth}{!}{
		\begin{tabular}{p{0.06\textwidth}*7{lS[table-format=1.4]@{$\ \pm\ $}S[table-format=1.4]}}
			\toprule
			 \rowcell{3}{ICD-9 \\ Task}	&& \multicolumn{11}{c}{Super Learner}	&& \multicolumn{8}{c}{Deep Learning}	\\	
			\cmidrule{3-13} \cmidrule{15-22}
			&& \multicolumn{2}{c}{Super Learner-I}	&& \multicolumn{2}{c}{Super Learner-II}	&& \multicolumn{2}{c}{Super Learner-II}	&& \multicolumn{2}{c}{Super Learner-II}	&& \multicolumn{2}{c}{FFN}	&& \multicolumn{2}{c}{RNN}	&& \multicolumn{2}{c}{MMDL} \\	
			&& \multicolumn{2}{c}{on Feature Set A}	&& \multicolumn{2}{c}{on Feature Set A}	&& \multicolumn{2}{c}{on Feature Set B}	&& \multicolumn{2}{c}{on Feature Set C}	&& \multicolumn{2}{c}{on Feature Set C}	&& \multicolumn{2}{c}{on Feature Set C}	&& \multicolumn{2}{c}{on Feature Set C} \\ \midrule	
			 1	&& 0.7371	& 0.0023	&& 0.7757	& 0.0049	&& 0.7635	& 0.0018	&& 0.7971	& 0.0044	&& 0.7643	& 0.0047	&& 0.7879	& 0.0133	&& \bfseries 0.8194	& \bfseries 0.0078 \\	
			 2	&& 0.8342	& 0.0091	&& 0.8530	& 0.0082	&& 0.8472	& 0.0056	&& 0.8770	& 0.0092	&& 0.8467	& 0.0062	&& 0.7651	& 0.0067	&& \bfseries 0.8912	& \bfseries 0.0154 \\	
			 3	&& 0.6835	& 0.0073	&& 0.7022	& 0.0070	&& 0.7108	& 0.0074	&& 0.7532	& 0.0061	&& 0.7203	& 0.0055	&& 0.7460	& 0.0059	&& \bfseries 0.7604	& \bfseries 0.0158 \\	
			 4	&& 0.6714	& 0.0052	&& 0.7048	& 0.0037	&& 0.7038	& 0.0023	&& 0.7814	& 0.0049	&& 0.7575	& 0.0067	&& 0.7831	& 0.0038	&& \bfseries 0.8048	& \bfseries 0.0131 \\	
			 5	&& 0.6540	& 0.0036	&& 0.6833	& 0.0047	&& 0.6825	& 0.0042	&& 0.7163	& 0.0036	&& 0.6680	& 0.0082	&& 0.6921	& 0.0032	&& \bfseries 0.7184	& \bfseries 0.0169 \\	
			 6	&& 0.6222	& 0.0028	&& 0.6505	& 0.0068	&& 0.6526	& 0.0027	&& 0.7107	& 0.0061	&& 0.6651	& 0.0071	&& 0.7109	& 0.0061	&& \bfseries 0.7387	& \bfseries 0.0122 \\	
			 7	&& 0.8270	& 0.0071	&& 0.8457	& 0.0067	&& 0.8424	& 0.0073	&& 0.8753	& 0.0060	&& 0.8435	& 0.0052	&& 0.8339	& 0.0068	&& \bfseries 0.8942	& \bfseries 0.0045 \\	
			 8	&& 0.6976	& 0.0049	&& 0.7340	& 0.0038	&& 0.7315	& 0.0066	&& 0.7636	& 0.0027	&& 0.7447	& 0.0029	&& 0.7957	& 0.0066	&& \bfseries 0.8148	& \bfseries 0.0093 \\	
			 9	&& 0.6518	& 0.0063	&& 0.6980	& 0.0093	&& 0.6993	& 0.0075	&& 0.7512	& 0.0107	&& 0.7023	& 0.0047	&& 0.7298	& 0.0102	&& \bfseries 0.7527	& \bfseries 0.0105 \\	
			 10	&& 0.7807	& 0.0024	&& 0.7978	& 0.0039	&& 0.7881	& 0.0041	&& 0.8266	& 0.0043	&& 0.8089	& 0.0016	&& 0.8354	& 0.0022	&& \bfseries 0.8512	& \bfseries 0.0105 \\	
			 11	&& 0.9454	& 0.0094	&& 0.9635	& 0.0060	&& 0.9592	& 0.0038	&& \bfseries 0.9692	& \bfseries 0.0081	&& 0.8200	& 0.0778	&& 0.8268	& 0.0322	&& 0.9035	& 0.0450 \\	
			 12	&& 0.6340	& 0.0151	&& 0.6578	& 0.0117	&& 0.6638	& 0.0161	&& 0.6969	& 0.0189	&& 0.6670	& 0.0068	&& 0.6554	& 0.0132	&& \bfseries 0.7378	& \bfseries 0.0141 \\	
			 13	&& 0.5908	& 0.0067	&& 0.6078	& 0.0049	&& 0.6141	& 0.0040	&& 0.6727	& 0.0055	&& 0.6185	& 0.0060	&& 0.6409	& 0.0088	&& \bfseries 0.6873	& \bfseries 0.0133 \\	
			 14	&& 0.7170	& 0.0172	&& 0.7172	& 0.0157	&& 0.7204	& 0.0211	&& 0.7166	& 0.0143	&& 0.7003	& 0.0128	&& 0.6191	& 0.0172	&& \bfseries 0.7346	& \bfseries 0.0251 \\	
			 15	&& 0.6507	& 0.0046	&& 0.6774	& 0.0048	&& 0.6603	& 0.0050	&& 0.6938	& 0.0028	&& 0.6583	& 0.0057	&& 0.6897	& 0.0063	&& \bfseries 0.7148	& \bfseries 0.0135 \\	
			 16	&& 0.5901	& 0.0156	&& 0.6200	& 0.0117	&& 0.6176	& 0.0144	&& 0.6615	& 0.0192	&& 0.6243	& 0.0121	&& 0.6300	& 0.0116	&& \bfseries 0.7011	& \bfseries 0.0207 \\	
			 17	&& 0.6769	& 0.0211	&& 0.6917	& 0.0094	&& 0.6991	& 0.0074	&& 0.7530	& 0.0236	&& 0.6976	& 0.0151	&& 0.6144	& 0.0223	&& \bfseries 0.8104	& \bfseries 0.0381 \\	
			 18	&& 0.6405	& 0.0076	&& 0.6683	& 0.0067	&& 0.6676	& 0.0066	&& 0.7050	& 0.0076	&& 0.6722	& 0.0115	&& 0.7007	& 0.0057	&& \bfseries 0.7319	& \bfseries 0.0240 \\	
			 19	&& 0.6385	& 0.0054	&& 0.6619	& 0.0068	&& 0.6762	& 0.0066	&& \bfseries 0.7358	& \bfseries 0.0052	&& 0.6931	& 0.0023	&& 0.7127	& 0.0019	&& 0.7348	& 0.0070 \\	
			 20	&& 0.6263	& 0.0071	&& 0.6538	& 0.0095	&& 0.6656	& 0.0043	&& 0.7175	& 0.0089	&& 0.6719	& 0.0053	&& 0.7108	& 0.0054	&& \bfseries 0.7426	& \bfseries 0.0136 \\ \cmidrule{1-22}	
			 Average	&& 0.6935	& 0.0080	&& 0.7182	& 0.0073	&& 0.7183	& 0.0069	&& 0.7587	& 0.0086	&& 0.7172	& 0.0104	&& 0.7240	& 0.0095	&& \bfseries 0.7772	& \bfseries 0.0165 \\	
			\bottomrule
		\end{tabular}
	}
\end{sidewaystable}

\subsubsection{Length of Stay Prediction Task Evaluation}
Table~\ref{table:los_mimic3} shows the performance measured by Mean Squared Error (MSE) of all methods on the task of forecasting length of stay task with first 24-hour data and first 48-hour data of MIMIC-III dataset. We observe that (i) all deep learning models such as FFN, RRN and MMDL trained on Feature set C outperform Super Learner models trained on Feature sets A,B, and C. (ii) MMDL model obtains best performance in terms of mean squared error (in hours), and significantly outperforms Super Learner II by nearly 50\%.

\subsubsection{Computation Time}
\label{sec:computation_time}
Our Python implementation of the Super Learner algorithm took about 25-30 mins for evaluating the in-hospital mortality task using Feature Set A, and it took about 3 hours for the Feature Set C. A deep Feed forward neural (FFN) network implemented using Keras took around 90 and 100 minutes for evaluating the same mortality task using Feature sets A and C respectively, while the MMDL model (shown in Figure~\ref{fig:ffn_lstm_struct}) took around 30 minutes and 1 hour for Feature sets A and C respectively. All our experiments was run on a 32-core Intel(R) Xeon(R) CPU E5-2630 v3 @ 2.40GHz machine with NVIDIA TITAN-X GPU processor.

\begin{table}[t]
    \renewcommand{\arraystretch}{1.2}
	\centering
	\caption{Length of stay task on MIMIC-III with first 24/48-hour data. Mean squared error (MSE) shown is in hours.}
	\label{table:los_mimic3}
	\resizebox{\textwidth}{!}{
		\begin{tabular}{ll*2{lS[table-format=5.4]@{$\ \pm\ $}S[table-format=4.4]}}
			\toprule
			\multicolumn{2}{c}{\textbf{Model and feature set}} && \multicolumn{2}{c}{\textbf{First 24-hour data}} && \multicolumn{2}{c}{\textbf{First 48-hour data}} \\ \midrule 
			\multirow{4}{*}{Super Learner} & Super Learner-I on Feature Set A &&  56420.6077 & 3739.0173 &&  58561.0081 & 4223.1785 \\
			& Super Learner-II on Feature Set A &&  54593.3317 & 3265.8292 &&  57454.7028 & 4349.3598 \\
			& Super Learner-II on Feature Set B && 55844.4209 & 3248.3224 && 54666.0875 & 4859.4577 \\
			& Super Learner-II on Feature Set C && 54608.1099 & 2923.9972 && 54400.5845 & 1582.4523 \\ \midrule
			\multirow{3}{*}{Deep Learning} & FFN on Feature Set C && 53410.0918 & 3207.9849 && 52642.6508  & 4373.4239 \\
			& RNN on Feature Set C &&  48702.7641 & 3768.5154 && 49556.8024 & 3794.0471 \\
			& MMDL on Feature Set C && \bfseries 36338.2015 & \bfseries 2672.3832 && \bfseries 36924.2312  & \bfseries 3566.4318 \\
			\bottomrule
		\end{tabular}
	}
\end{table}

	\section{Summary}
	\label{sec:summary}
	In this paper, we presented exhaustive benchmarking evaluation results of deep learning models, several machine learning models and ICU scoring systems on various clinical prediction tasks using the publicly available MIMIC-III datasets. We demonstrated that deep learning models consistently outperform all the other approaches especially when a large number of raw clinical time series data is used as input features to the prediction models.


\subsubsection*{Acknowledgments}
This material is based upon work supported by the NSF research grants IIS-1134990, IIS-1254206, and Samsung GRO Grant. Any opinions, findings, and conclusions or recommendations expressed in this material are those of the author(s) and do not necessarily reflect the views of the funding agencies.

    \bibliographystyle{elsarticle-num-names}
	\bibliography{references}

\begin{thebibliography}{63}
\expandafter\ifx\csname natexlab\endcsname\relax\def\natexlab#1{#1}\fi
\providecommand{\url}[1]{\texttt{#1}}
\providecommand{\href}[2]{#2}
\providecommand{\path}[1]{#1}
\providecommand{\DOIprefix}{doi:}
\providecommand{\ArXivprefix}{arXiv:}
\providecommand{\URLprefix}{URL: }
\providecommand{\Pubmedprefix}{pmid:}
\providecommand{\doi}[1]{\href{http://dx.doi.org/#1}{\path{#1}}}
\providecommand{\Pubmed}[1]{\href{pmid:#1}{\path{#1}}}
\providecommand{\bibinfo}[2]{#2}
\ifx\xfnm\relax \def\xfnm[#1]{\unskip,\space#1}\fi
\bibitem[{Che et~al.(2016)Che, Purushotham, Cho, Sontag, and Liu}]{rel4}
\bibinfo{author}{Z.~Che}, \bibinfo{author}{S.~Purushotham},
  \bibinfo{author}{K.~Cho}, \bibinfo{author}{D.~Sontag},
  \bibinfo{author}{Y.~Liu},
\newblock \bibinfo{title}{Recurrent neural networks for multivariate time
  series with missing values},
\newblock \bibinfo{journal}{arXiv preprint arXiv:1606.01865}
  (\bibinfo{year}{2016}).
\bibitem[{Harutyunyan et~al.(2017)Harutyunyan, Khachatrian, Kale, and
  Galstyan}]{harutyunyan2017multitask}
\bibinfo{author}{H.~Harutyunyan}, \bibinfo{author}{H.~Khachatrian},
  \bibinfo{author}{D.~C. Kale}, \bibinfo{author}{A.~Galstyan},
\newblock \bibinfo{title}{Multitask learning and benchmarking with clinical
  time series data},
\newblock \bibinfo{journal}{arXiv preprint arXiv:1703.07771}
  (\bibinfo{year}{2017}).
\bibitem[{Le~Gall et~al.(1993)Le~Gall, Lemeshow, and Saulnier}]{le1993new}
\bibinfo{author}{J.-R. Le~Gall}, \bibinfo{author}{S.~Lemeshow},
  \bibinfo{author}{F.~Saulnier},
\newblock \bibinfo{title}{A new simplified acute physiology score (saps ii)
  based on a european/north american multicenter study},
\newblock \bibinfo{journal}{Jama} \bibinfo{volume}{270} (\bibinfo{year}{1993})
  \bibinfo{pages}{2957--2963}.
\bibitem[{Lee et~al.(2011)Lee, Scott, Villarroel, Clifford, Saeed, and
  Mark}]{lee2011open}
\bibinfo{author}{J.~Lee}, \bibinfo{author}{D.~J. Scott},
  \bibinfo{author}{M.~Villarroel}, \bibinfo{author}{G.~D. Clifford},
  \bibinfo{author}{M.~Saeed}, \bibinfo{author}{R.~G. Mark},
\newblock \bibinfo{title}{Open-access mimic-ii database for intensive care
  research},
\newblock in: \bibinfo{booktitle}{Engineering in Medicine and Biology Society,
  EMBC, 2011 Annual International Conference of the IEEE},
  \bibinfo{organization}{IEEE}, \bibinfo{year}{2011}, pp.
  \bibinfo{pages}{8315--8318}.
\bibitem[{Johnson et~al.(2016)Johnson, Pollard, Shen, Lehman, Feng, Ghassemi,
  Moody, Szolovits, Celi, and Mark}]{johnson2016mimic}
\bibinfo{author}{A.~E. Johnson}, \bibinfo{author}{T.~J. Pollard},
  \bibinfo{author}{L.~Shen}, \bibinfo{author}{L.-w.~H. Lehman},
  \bibinfo{author}{M.~Feng}, \bibinfo{author}{M.~Ghassemi},
  \bibinfo{author}{B.~Moody}, \bibinfo{author}{P.~Szolovits},
  \bibinfo{author}{L.~A. Celi}, \bibinfo{author}{R.~G. Mark},
\newblock \bibinfo{title}{Mimic-iii, a freely accessible critical care
  database},
\newblock \bibinfo{journal}{Scientific data} \bibinfo{volume}{3}
  (\bibinfo{year}{2016}).
\bibitem[{Johnson et~al.(2017)Johnson, Pollard, and
  Mark}]{johnsonreproducibility}
\bibinfo{author}{A.~E. Johnson}, \bibinfo{author}{T.~J. Pollard},
  \bibinfo{author}{R.~G. Mark},
\newblock \bibinfo{title}{Reproducibility in critical care: a mortality
  prediction case study},
\newblock \bibinfo{journal}{Machine Learning for Healthcare (MLHC)}
  (\bibinfo{year}{2017}).
\bibitem[{Che et~al.(2015)Che, Kale, Li, Bahadori, and Liu}]{che2015deep}
\bibinfo{author}{Z.~Che}, \bibinfo{author}{D.~Kale}, \bibinfo{author}{W.~Li},
  \bibinfo{author}{M.~T. Bahadori}, \bibinfo{author}{Y.~Liu},
\newblock \bibinfo{title}{Deep computational phenotyping},
\newblock in: \bibinfo{booktitle}{Proceedings of the 21th ACM SIGKDD
  International Conference on Knowledge Discovery and Data Mining},
  \bibinfo{organization}{ACM}, \bibinfo{year}{2015}, pp.
  \bibinfo{pages}{507--516}.
\bibitem[{Che et~al.(2016)Che, Purushotham, Khemani, and
  Liu}]{che2016interpretable}
\bibinfo{author}{Z.~Che}, \bibinfo{author}{S.~Purushotham},
  \bibinfo{author}{R.~Khemani}, \bibinfo{author}{Y.~Liu},
\newblock \bibinfo{title}{Interpretable deep models for icu outcome
  prediction},
\newblock in: \bibinfo{booktitle}{AMIA Annual Symposium Proceedings}, volume
  \bibinfo{volume}{2016}, \bibinfo{organization}{American Medical Informatics
  Association}, \bibinfo{year}{2016}, p. \bibinfo{pages}{371}.
\bibitem[{Caballero~Barajas and Akella(2015)}]{rel1}
\bibinfo{author}{K.~L. Caballero~Barajas}, \bibinfo{author}{R.~Akella},
\newblock \bibinfo{title}{Dynamically modeling patient's health state from
  electronic medical records: a time series approach},
\newblock in: \bibinfo{booktitle}{Proceedings of the 21th ACM SIGKDD
  International Conference on Knowledge Discovery and Data Mining},
  \bibinfo{organization}{ACM}, \bibinfo{year}{2015}, pp.
  \bibinfo{pages}{69--78}.
\bibitem[{Calvert et~al.(2016)Calvert, Mao, Rogers, Barton, Jay, Desautels,
  Mohamadlou, Jan, and Das}]{rel2}
\bibinfo{author}{J.~Calvert}, \bibinfo{author}{Q.~Mao}, \bibinfo{author}{A.~J.
  Rogers}, \bibinfo{author}{C.~Barton}, \bibinfo{author}{M.~Jay},
  \bibinfo{author}{T.~Desautels}, \bibinfo{author}{H.~Mohamadlou},
  \bibinfo{author}{J.~Jan}, \bibinfo{author}{R.~Das},
\newblock \bibinfo{title}{A computational approach to mortality prediction of
  alcohol use disorder inpatients},
\newblock \bibinfo{journal}{Computers in biology and medicine}
  \bibinfo{volume}{75} (\bibinfo{year}{2016}) \bibinfo{pages}{74--79}.
\bibitem[{Celi et~al.(2012)Celi, Galvin, Davidzon, Lee, Scott, and Mark}]{rel3}
\bibinfo{author}{L.~A. Celi}, \bibinfo{author}{S.~Galvin},
  \bibinfo{author}{G.~Davidzon}, \bibinfo{author}{J.~Lee},
  \bibinfo{author}{D.~Scott}, \bibinfo{author}{R.~Mark},
\newblock \bibinfo{title}{A database-driven decision support system: customized
  mortality prediction},
\newblock \bibinfo{journal}{Journal of personalized medicine}
  \bibinfo{volume}{2} (\bibinfo{year}{2012}) \bibinfo{pages}{138--148}.
\bibitem[{Ghassemi et~al.(2015)Ghassemi, Pimentel, Naumann, Brennan, Clifton,
  Szolovits, and Feng}]{rel5}
\bibinfo{author}{M.~Ghassemi}, \bibinfo{author}{M.~A. Pimentel},
  \bibinfo{author}{T.~Naumann}, \bibinfo{author}{T.~Brennan},
  \bibinfo{author}{D.~A. Clifton}, \bibinfo{author}{P.~Szolovits},
  \bibinfo{author}{M.~Feng},
\newblock \bibinfo{title}{A multivariate timeseries modeling approach to
  severity of illness assessment and forecasting in icu with sparse,
  heterogeneous clinical data.},
\newblock in: \bibinfo{booktitle}{AAAI}, \bibinfo{year}{2015}, pp.
  \bibinfo{pages}{446--453}.
\bibitem[{Ghassemi et~al.(2014)Ghassemi, Naumann, Doshi-Velez, Brimmer, Joshi,
  Rumshisky, and Szolovits}]{rel6}
\bibinfo{author}{M.~Ghassemi}, \bibinfo{author}{T.~Naumann},
  \bibinfo{author}{F.~Doshi-Velez}, \bibinfo{author}{N.~Brimmer},
  \bibinfo{author}{R.~Joshi}, \bibinfo{author}{A.~Rumshisky},
  \bibinfo{author}{P.~Szolovits},
\newblock \bibinfo{title}{Unfolding physiological state: Mortality modelling in
  intensive care units},
\newblock in: \bibinfo{booktitle}{Proceedings of the 20th ACM SIGKDD
  international conference on Knowledge discovery and data mining},
  \bibinfo{organization}{ACM}, \bibinfo{year}{2014}, pp.
  \bibinfo{pages}{75--84}.
\bibitem[{Hoogendoorn et~al.(2016)Hoogendoorn, el~Hassouni, Mok, Ghassemi, and
  Szolovits}]{rel7}
\bibinfo{author}{M.~Hoogendoorn}, \bibinfo{author}{A.~el~Hassouni},
  \bibinfo{author}{K.~Mok}, \bibinfo{author}{M.~Ghassemi},
  \bibinfo{author}{P.~Szolovits},
\newblock \bibinfo{title}{Prediction using patient comparison vs. modeling: A
  case study for mortality prediction},
\newblock in: \bibinfo{booktitle}{Engineering in Medicine and Biology Society
  (EMBC), 2016 IEEE 38th Annual International Conference of the},
  \bibinfo{organization}{IEEE}, \bibinfo{year}{2016}, pp.
  \bibinfo{pages}{2464--2467}.
\bibitem[{Joshi and Szolovits(2012)}]{rel8}
\bibinfo{author}{R.~Joshi}, \bibinfo{author}{P.~Szolovits},
\newblock \bibinfo{title}{Prognostic physiology: modeling patient severity in
  intensive care units using radial domain folding},
\newblock in: \bibinfo{booktitle}{AMIA Annual Symposium Proceedings}, volume
  \bibinfo{volume}{2012}, \bibinfo{organization}{American Medical Informatics
  Association}, \bibinfo{year}{2012}, p. \bibinfo{pages}{1276}.
\bibitem[{Lee and Maslove(2017)}]{rel9}
\bibinfo{author}{J.~Lee}, \bibinfo{author}{D.~M. Maslove},
\newblock \bibinfo{title}{Customization of a severity of illness score using
  local electronic medical record data},
\newblock \bibinfo{journal}{Journal of intensive care medicine}
  \bibinfo{volume}{32} (\bibinfo{year}{2017}) \bibinfo{pages}{38--47}.
\bibitem[{Lehman et~al.(2012)Lehman, Saeed, Long, Lee, and Mark}]{rel10}
\bibinfo{author}{L.-w. Lehman}, \bibinfo{author}{M.~Saeed},
  \bibinfo{author}{W.~Long}, \bibinfo{author}{J.~Lee},
  \bibinfo{author}{R.~Mark},
\newblock \bibinfo{title}{Risk stratification of icu patients using topic
  models inferred from unstructured progress notes},
\newblock in: \bibinfo{booktitle}{AMIA annual symposium proceedings}, volume
  \bibinfo{volume}{2012}, \bibinfo{organization}{American Medical Informatics
  Association}, \bibinfo{year}{2012}, p. \bibinfo{pages}{505}.
\bibitem[{Luo et~al.(2016)Luo, Xin, Joshi, Celi, and Szolovits}]{rel11}
\bibinfo{author}{Y.~Luo}, \bibinfo{author}{Y.~Xin}, \bibinfo{author}{R.~Joshi},
  \bibinfo{author}{L.~A. Celi}, \bibinfo{author}{P.~Szolovits},
\newblock \bibinfo{title}{Predicting icu mortality risk by grouping temporal
  trends from a multivariate panel of physiologic measurements.},
\newblock in: \bibinfo{booktitle}{AAAI}, \bibinfo{year}{2016}, pp.
  \bibinfo{pages}{42--50}.
\bibitem[{Joshi et~al.(2016)Joshi, Gunasekar, Sontag, and Joydeep}]{rel12}
\bibinfo{author}{S.~Joshi}, \bibinfo{author}{S.~Gunasekar},
  \bibinfo{author}{D.~Sontag}, \bibinfo{author}{G.~Joydeep},
\newblock \bibinfo{title}{Identifiable phenotyping using constrained
  non-negative matrix factorization},
\newblock in: \bibinfo{booktitle}{Machine Learning for Healthcare Conference},
  \bibinfo{year}{2016}, pp. \bibinfo{pages}{17--41}.
\bibitem[{Lee et~al.(2015)Lee, Maslove, and Dubin}]{rel13}
\bibinfo{author}{J.~Lee}, \bibinfo{author}{D.~M. Maslove},
  \bibinfo{author}{J.~A. Dubin},
\newblock \bibinfo{title}{Personalized mortality prediction driven by
  electronic medical data and a patient similarity metric},
\newblock \bibinfo{journal}{PloS one} \bibinfo{volume}{10}
  (\bibinfo{year}{2015}) \bibinfo{pages}{e0127428}.
\bibitem[{Lee(2017)}]{rel14}
\bibinfo{author}{J.~Lee},
\newblock \bibinfo{title}{Patient-specific predictive modeling using random
  forests: An observational study for the critically ill},
\newblock \bibinfo{journal}{JMIR medical informatics} \bibinfo{volume}{5}
  (\bibinfo{year}{2017}).
\bibitem[{Luo and Rumshisky(2016)}]{rel15}
\bibinfo{author}{Y.-F. Luo}, \bibinfo{author}{A.~Rumshisky},
\newblock \bibinfo{title}{Interpretable topic features for post-icu mortality
  prediction},
\newblock in: \bibinfo{booktitle}{AMIA Annual Symposium Proceedings}, volume
  \bibinfo{volume}{2016}, \bibinfo{organization}{American Medical Informatics
  Association}, \bibinfo{year}{2016}, p. \bibinfo{pages}{827}.
\bibitem[{Vincent et~al.(1996)Vincent, Moreno, Takala, Willatts,
  De~Mendon{\c{c}}a, Bruining, Reinhart, Suter, and Thijs}]{vincent1996sofa}
\bibinfo{author}{J.-L. Vincent}, \bibinfo{author}{R.~Moreno},
  \bibinfo{author}{J.~Takala}, \bibinfo{author}{S.~Willatts},
  \bibinfo{author}{A.~De~Mendon{\c{c}}a}, \bibinfo{author}{H.~Bruining},
  \bibinfo{author}{C.~Reinhart}, \bibinfo{author}{P.~Suter},
  \bibinfo{author}{L.~Thijs},
\newblock \bibinfo{title}{The sofa (sepsis-related organ failure assessment)
  score to describe organ dysfunction/failure},
\newblock \bibinfo{journal}{Intensive care medicine} \bibinfo{volume}{22}
  (\bibinfo{year}{1996}) \bibinfo{pages}{707--710}.
\bibitem[{Knaus et~al.(1981)Knaus, Zimmerman, Wagner, Draper, and
  Lawrence}]{knaus1981apache}
\bibinfo{author}{W.~A. Knaus}, \bibinfo{author}{J.~E. Zimmerman},
  \bibinfo{author}{D.~P. Wagner}, \bibinfo{author}{E.~A. Draper},
  \bibinfo{author}{D.~E. Lawrence},
\newblock \bibinfo{title}{Apache-acute physiology and chronic health
  evaluation: a physiologically based classification system.},
\newblock \bibinfo{journal}{Critical care medicine} \bibinfo{volume}{9}
  (\bibinfo{year}{1981}) \bibinfo{pages}{591--597}.
\bibitem[{Dybowski et~al.(1996)Dybowski, Gant, Weller, and
  Chang}]{dybowski1996prediction}
\bibinfo{author}{R.~Dybowski}, \bibinfo{author}{V.~Gant},
  \bibinfo{author}{P.~Weller}, \bibinfo{author}{R.~Chang},
\newblock \bibinfo{title}{Prediction of outcome in critically ill patients
  using artificial neural network synthesised by genetic algorithm},
\newblock \bibinfo{journal}{The Lancet} \bibinfo{volume}{347}
  (\bibinfo{year}{1996}) \bibinfo{pages}{1146--1150}.
\bibitem[{Kim et~al.(2011)Kim, Kim, and Park}]{kim2011comparison}
\bibinfo{author}{S.~Kim}, \bibinfo{author}{W.~Kim}, \bibinfo{author}{R.~W.
  Park},
\newblock \bibinfo{title}{A comparison of intensive care unit mortality
  prediction models through the use of data mining techniques},
\newblock \bibinfo{journal}{Healthcare informatics research}
  \bibinfo{volume}{17} (\bibinfo{year}{2011}) \bibinfo{pages}{232--243}.
\bibitem[{Tu and Guerriere(1993)}]{tu1993use}
\bibinfo{author}{J.~V. Tu}, \bibinfo{author}{M.~R. Guerriere},
\newblock \bibinfo{title}{Use of a neural network as a predictive instrument
  for length of stay in the intensive care unit following cardiac surgery},
\newblock \bibinfo{journal}{Computers and biomedical research}
  \bibinfo{volume}{26} (\bibinfo{year}{1993}) \bibinfo{pages}{220--229}.
\bibitem[{Doig et~al.(1993)Doig, Inman, Sibbald, Martin, and
  Robertson}]{doig1993modeling}
\bibinfo{author}{G.~Doig}, \bibinfo{author}{K.~Inman},
  \bibinfo{author}{W.~Sibbald}, \bibinfo{author}{C.~Martin},
  \bibinfo{author}{J.~Robertson},
\newblock \bibinfo{title}{Modeling mortality in the intensive care unit:
  comparing the performance of a back-propagation, associative-learning neural
  network with multivariate logistic regression.},
\newblock in: \bibinfo{booktitle}{Proceedings of the Annual Symposium on
  Computer Application in Medical Care}, \bibinfo{organization}{American
  Medical Informatics Association}, \bibinfo{year}{1993}, p.
  \bibinfo{pages}{361}.
\bibitem[{Hanson~III and Marshall(2001)}]{hanson2001artificial}
\bibinfo{author}{C.~W. Hanson~III}, \bibinfo{author}{B.~E. Marshall},
\newblock \bibinfo{title}{Artificial intelligence applications in the intensive
  care unit},
\newblock \bibinfo{journal}{Critical care medicine} \bibinfo{volume}{29}
  (\bibinfo{year}{2001}) \bibinfo{pages}{427--435}.
\bibitem[{Pirracchio(2016)}]{pirracchio2016mortality}
\bibinfo{author}{R.~Pirracchio},
\newblock \bibinfo{title}{Mortality prediction in the icu based on mimic-ii
  results from the super icu learner algorithm (sicula) project},
\newblock in: \bibinfo{booktitle}{Secondary Analysis of Electronic Health
  Records}, \bibinfo{publisher}{Springer}, \bibinfo{year}{2016}, pp.
  \bibinfo{pages}{295--313}.
\bibitem[{Polley and Van~der Laan(2010)}]{polley2010super}
\bibinfo{author}{E.~C. Polley}, \bibinfo{author}{M.~J. Van~der Laan},
\newblock \bibinfo{title}{Super learner in prediction},
\newblock \bibinfo{journal}{U.C. Berkeley Division of Biostatistics Working
  Paper Series}  (\bibinfo{year}{2010}).
\bibitem[{Caruana et~al.(1996)Caruana, Baluja, and Mitchell}]{caruana1996using}
\bibinfo{author}{R.~Caruana}, \bibinfo{author}{S.~Baluja},
  \bibinfo{author}{T.~Mitchell},
\newblock \bibinfo{title}{Using the future to" sort out" the present: Rankprop
  and multitask learning for medical risk evaluation},
\newblock in: \bibinfo{booktitle}{Advances in neural information processing
  systems}, \bibinfo{year}{1996}, pp. \bibinfo{pages}{959--965}.
\bibitem[{Cooper et~al.(1997)Cooper, Aliferis, Ambrosino, Aronis, Buchanan,
  Caruana, Fine, Glymour, Gordon, Hanusa et~al.}]{cooper1997evaluation}
\bibinfo{author}{G.~F. Cooper}, \bibinfo{author}{C.~F. Aliferis},
  \bibinfo{author}{R.~Ambrosino}, \bibinfo{author}{J.~Aronis},
  \bibinfo{author}{B.~G. Buchanan}, \bibinfo{author}{R.~Caruana},
  \bibinfo{author}{M.~J. Fine}, \bibinfo{author}{C.~Glymour},
  \bibinfo{author}{G.~Gordon}, \bibinfo{author}{B.~H. Hanusa}, et~al.,
\newblock \bibinfo{title}{An evaluation of machine-learning methods for
  predicting pneumonia mortality},
\newblock \bibinfo{journal}{Artificial intelligence in medicine}
  \bibinfo{volume}{9} (\bibinfo{year}{1997}) \bibinfo{pages}{107--138}.
\bibitem[{Lasko et~al.(2013)Lasko, Denny, and Levy}]{lasko2013computational}
\bibinfo{author}{T.~A. Lasko}, \bibinfo{author}{J.~C. Denny},
  \bibinfo{author}{M.~A. Levy},
\newblock \bibinfo{title}{Computational phenotype discovery using unsupervised
  feature learning over noisy, sparse, and irregular clinical data},
\newblock \bibinfo{journal}{PloS one} \bibinfo{volume}{8}
  (\bibinfo{year}{2013}) \bibinfo{pages}{e66341}.
\bibitem[{Oellrich et~al.(2015)Oellrich, Collier, Groza, Rebholz-Schuhmann,
  Shah, Bodenreider, Boland, Georgiev, Liu, Livingston
  et~al.}]{oellrich2015digital}
\bibinfo{author}{A.~Oellrich}, \bibinfo{author}{N.~Collier},
  \bibinfo{author}{T.~Groza}, \bibinfo{author}{D.~Rebholz-Schuhmann},
  \bibinfo{author}{N.~Shah}, \bibinfo{author}{O.~Bodenreider},
  \bibinfo{author}{M.~R. Boland}, \bibinfo{author}{I.~Georgiev},
  \bibinfo{author}{H.~Liu}, \bibinfo{author}{K.~Livingston}, et~al.,
\newblock \bibinfo{title}{The digital revolution in phenotyping},
\newblock \bibinfo{journal}{Briefings in bioinformatics}
  (\bibinfo{year}{2015}) \bibinfo{pages}{bbv083}.
\bibitem[{Che et~al.(2015)Che, Purushotham, Khemani, and
  Liu}]{che2015distilling}
\bibinfo{author}{Z.~Che}, \bibinfo{author}{S.~Purushotham},
  \bibinfo{author}{R.~Khemani}, \bibinfo{author}{Y.~Liu},
\newblock \bibinfo{title}{Distilling knowledge from deep networks with
  applications to healthcare domain},
\newblock \bibinfo{journal}{arXiv preprint arXiv:1512.03542}
  (\bibinfo{year}{2015}).
\bibitem[{Dabek and Caban(2015)}]{dabek2015neural}
\bibinfo{author}{F.~Dabek}, \bibinfo{author}{J.~J. Caban},
\newblock \bibinfo{title}{A neural network based model for predicting
  psychological conditions},
\newblock in: \bibinfo{booktitle}{Brain Informatics and Health},
  \bibinfo{publisher}{Springer}, \bibinfo{year}{2015}, pp.
  \bibinfo{pages}{252--261}.
\bibitem[{Hammerla et~al.(2015)Hammerla, Fisher, Andras, Rochester, Walker, and
  Pl{\"o}tz}]{hammerla2015pd}
\bibinfo{author}{N.~Y. Hammerla}, \bibinfo{author}{J.~M. Fisher},
  \bibinfo{author}{P.~Andras}, \bibinfo{author}{L.~Rochester},
  \bibinfo{author}{R.~Walker}, \bibinfo{author}{T.~Pl{\"o}tz},
\newblock \bibinfo{title}{Pd disease state assessment in naturalistic
  environments using deep learning},
\newblock in: \bibinfo{booktitle}{Twenty-Ninth AAAI Conference on Artificial
  Intelligence}, \bibinfo{year}{2015}, pp. \bibinfo{pages}{1742--1748}.
\bibitem[{Lipton et~al.(2015)Lipton, Kale, Elkan, and
  Wetzell}]{lipton2015learning}
\bibinfo{author}{Z.~C. Lipton}, \bibinfo{author}{D.~C. Kale},
  \bibinfo{author}{C.~Elkan}, \bibinfo{author}{R.~Wetzell},
\newblock \bibinfo{title}{Learning to diagnose with lstm recurrent neural
  networks},
\newblock \bibinfo{journal}{arXiv preprint arXiv:1511.03677}
  (\bibinfo{year}{2015}).
\bibitem[{Purushotham et~al.(2017)Purushotham, Carvalho, Nilanon, and
  Liu}]{purushotham2016variational}
\bibinfo{author}{S.~Purushotham}, \bibinfo{author}{W.~Carvalho},
  \bibinfo{author}{T.~Nilanon}, \bibinfo{author}{Y.~Liu},
\newblock \bibinfo{title}{Variational recurrent adversarial deep domain
  adaptation},
\newblock \bibinfo{journal}{International Conference on Learning
  Representations (ICLR)}  (\bibinfo{year}{2017}).
\bibitem[{Hochreiter and Schmidhuber(1997)}]{hochreiter1997long}
\bibinfo{author}{S.~Hochreiter}, \bibinfo{author}{J.~Schmidhuber},
\newblock \bibinfo{title}{Long short-term memory},
\newblock \bibinfo{journal}{Neural computation} \bibinfo{volume}{9}
  (\bibinfo{year}{1997}) \bibinfo{pages}{1735--1780}.
\bibitem[{Silva et~al.(2012)Silva, Moody, Scott, Celi, and
  Mark}]{silva2012predicting}
\bibinfo{author}{I.~Silva}, \bibinfo{author}{G.~Moody}, \bibinfo{author}{D.~J.
  Scott}, \bibinfo{author}{L.~A. Celi}, \bibinfo{author}{R.~G. Mark},
\newblock \bibinfo{title}{Predicting in-hospital mortality of icu patients: The
  physionet/computing in cardiology challenge 2012},
\newblock in: \bibinfo{booktitle}{Computing in Cardiology (CinC), 2012},
  \bibinfo{organization}{IEEE}, \bibinfo{year}{2012}, pp.
  \bibinfo{pages}{245--248}.
\bibitem[{Van~der Laan et~al.(2007)Van~der Laan, Polley, and
  Hubbard}]{van2007super}
\bibinfo{author}{M.~J. Van~der Laan}, \bibinfo{author}{E.~C. Polley},
  \bibinfo{author}{A.~E. Hubbard},
\newblock \bibinfo{title}{Super learner},
\newblock \bibinfo{journal}{Statistical applications in genetics and molecular
  biology} \bibinfo{volume}{6} (\bibinfo{year}{2007}).
\bibitem[{LeCun et~al.(2015)LeCun, Bengio, and Hinton}]{lecun2015deep}
\bibinfo{author}{Y.~LeCun}, \bibinfo{author}{Y.~Bengio},
  \bibinfo{author}{G.~Hinton},
\newblock \bibinfo{title}{Deep learning},
\newblock \bibinfo{journal}{Nature} \bibinfo{volume}{521}
  (\bibinfo{year}{2015}) \bibinfo{pages}{436--444}.
\bibitem[{Larochelle et~al.(2009)Larochelle, Bengio, Louradour, and
  Lamblin}]{larochelle2009exploring}
\bibinfo{author}{H.~Larochelle}, \bibinfo{author}{Y.~Bengio},
  \bibinfo{author}{J.~Louradour}, \bibinfo{author}{P.~Lamblin},
\newblock \bibinfo{title}{Exploring strategies for training deep neural
  networks},
\newblock \bibinfo{journal}{Journal of Machine Learning Research}
  \bibinfo{volume}{10} (\bibinfo{year}{2009}) \bibinfo{pages}{1--40}.
\bibitem[{Bengio et~al.(2013)Bengio, Courville, and
  Vincent}]{bengio2013representation}
\bibinfo{author}{Y.~Bengio}, \bibinfo{author}{A.~Courville},
  \bibinfo{author}{P.~Vincent},
\newblock \bibinfo{title}{Representation learning: A review and new
  perspectives},
\newblock \bibinfo{journal}{Pattern Analysis and Machine Intelligence, IEEE
  Transactions on} \bibinfo{volume}{35} (\bibinfo{year}{2013})
  \bibinfo{pages}{1798--1828}.
\bibitem[{Dahl et~al.(2010)Dahl, Mohamed, Hinton et~al.}]{dahl2010phone}
\bibinfo{author}{G.~Dahl}, \bibinfo{author}{A.-r. Mohamed},
  \bibinfo{author}{G.~E. Hinton}, et~al.,
\newblock \bibinfo{title}{Phone recognition with the mean-covariance restricted
  boltzmann machine},
\newblock in: \bibinfo{booktitle}{Advances in neural information processing
  systems}, \bibinfo{year}{2010}, pp. \bibinfo{pages}{469--477}.
\bibitem[{Dahl et~al.(2012)Dahl, Yu, Deng, and Acero}]{dahl2012context}
\bibinfo{author}{G.~E. Dahl}, \bibinfo{author}{D.~Yu},
  \bibinfo{author}{L.~Deng}, \bibinfo{author}{A.~Acero},
\newblock \bibinfo{title}{Context-dependent pre-trained deep neural networks
  for large-vocabulary speech recognition},
\newblock \bibinfo{journal}{IEEE Transactions on audio, speech, and language
  processing} \bibinfo{volume}{20} (\bibinfo{year}{2012})
  \bibinfo{pages}{30--42}.
\bibitem[{Krizhevsky et~al.(2012)Krizhevsky, Sutskever, and
  Hinton}]{krizhevsky2012imagenet}
\bibinfo{author}{A.~Krizhevsky}, \bibinfo{author}{I.~Sutskever},
  \bibinfo{author}{G.~E. Hinton},
\newblock \bibinfo{title}{Imagenet classification with deep convolutional
  neural networks},
\newblock in: \bibinfo{booktitle}{Advances in neural information processing
  systems}, \bibinfo{year}{2012}, pp. \bibinfo{pages}{1097--1105}.
\bibitem[{Jia et~al.(2014)Jia, Shelhamer, Donahue, Karayev, Long, Girshick,
  Guadarrama, and Darrell}]{jia2014caffe}
\bibinfo{author}{Y.~Jia}, \bibinfo{author}{E.~Shelhamer},
  \bibinfo{author}{J.~Donahue}, \bibinfo{author}{S.~Karayev},
  \bibinfo{author}{J.~Long}, \bibinfo{author}{R.~Girshick},
  \bibinfo{author}{S.~Guadarrama}, \bibinfo{author}{T.~Darrell},
\newblock \bibinfo{title}{Caffe: Convolutional architecture for fast feature
  embedding},
\newblock in: \bibinfo{booktitle}{Proceedings of the 22nd ACM international
  conference on Multimedia}, \bibinfo{organization}{ACM}, \bibinfo{year}{2014},
  pp. \bibinfo{pages}{675--678}.
\bibitem[{Szegedy et~al.(2015)Szegedy, Liu, Jia, Sermanet, Reed, Anguelov,
  Erhan, Vanhoucke, and Rabinovich}]{szegedy2015going}
\bibinfo{author}{C.~Szegedy}, \bibinfo{author}{W.~Liu},
  \bibinfo{author}{Y.~Jia}, \bibinfo{author}{P.~Sermanet},
  \bibinfo{author}{S.~Reed}, \bibinfo{author}{D.~Anguelov},
  \bibinfo{author}{D.~Erhan}, \bibinfo{author}{V.~Vanhoucke},
  \bibinfo{author}{A.~Rabinovich},
\newblock \bibinfo{title}{Going deeper with convolutions},
\newblock in: \bibinfo{booktitle}{Proceedings of the IEEE conference on
  computer vision and pattern recognition}, \bibinfo{year}{2015}, pp.
  \bibinfo{pages}{1--9}.
\bibitem[{Mikolov et~al.(2011)Mikolov, Deoras, Kombrink, Burget, and
  {\v{C}}ernock{\`y}}]{mikolov2011empirical}
\bibinfo{author}{T.~Mikolov}, \bibinfo{author}{A.~Deoras},
  \bibinfo{author}{S.~Kombrink}, \bibinfo{author}{L.~Burget},
  \bibinfo{author}{J.~{\v{C}}ernock{\`y}},
\newblock \bibinfo{title}{Empirical evaluation and combination of advanced
  language modeling techniques},
\newblock in: \bibinfo{booktitle}{Twelfth Annual Conference of the
  International Speech Communication Association}, \bibinfo{year}{2011}, pp.
  \bibinfo{pages}{605--608}.
\bibitem[{Bordes et~al.(2012)Bordes, Glorot, Weston, and
  Bengio}]{bordes2012joint}
\bibinfo{author}{A.~Bordes}, \bibinfo{author}{X.~Glorot},
  \bibinfo{author}{J.~Weston}, \bibinfo{author}{Y.~Bengio},
\newblock \bibinfo{title}{Joint learning of words and meaning representations
  for open-text semantic parsing},
\newblock in: \bibinfo{booktitle}{Artificial Intelligence and Statistics},
  \bibinfo{year}{2012}, pp. \bibinfo{pages}{127--135}.
\bibitem[{Mikolov et~al.(2013{\natexlab{a}})Mikolov, Yih, and
  Zweig}]{mikolov2013linguistic}
\bibinfo{author}{T.~Mikolov}, \bibinfo{author}{W.-t. Yih},
  \bibinfo{author}{G.~Zweig},
\newblock \bibinfo{title}{Linguistic regularities in continuous space word
  representations.},
\newblock in: \bibinfo{booktitle}{hlt-Naacl}, volume~\bibinfo{volume}{13},
  \bibinfo{year}{2013}{\natexlab{a}}, pp. \bibinfo{pages}{746--751}.
\bibitem[{Mikolov et~al.(2013{\natexlab{b}})Mikolov, Sutskever, Chen, Corrado,
  and Dean}]{mikolov2013distributed}
\bibinfo{author}{T.~Mikolov}, \bibinfo{author}{I.~Sutskever},
  \bibinfo{author}{K.~Chen}, \bibinfo{author}{G.~S. Corrado},
  \bibinfo{author}{J.~Dean},
\newblock \bibinfo{title}{Distributed representations of words and phrases and
  their compositionality},
\newblock in: \bibinfo{booktitle}{Advances in neural information processing
  systems}, \bibinfo{year}{2013}{\natexlab{b}}, pp.
  \bibinfo{pages}{3111--3119}.
\bibitem[{Hornik et~al.(1989)Hornik, Stinchcombe, and
  White}]{hornik1989multilayer}
\bibinfo{author}{K.~Hornik}, \bibinfo{author}{M.~Stinchcombe},
  \bibinfo{author}{H.~White},
\newblock \bibinfo{title}{Multilayer feedforward networks are universal
  approximators},
\newblock \bibinfo{journal}{Neural networks} \bibinfo{volume}{2}
  (\bibinfo{year}{1989}) \bibinfo{pages}{359--366}.
\bibitem[{Nair and Hinton(2010)}]{nair2010rectified}
\bibinfo{author}{V.~Nair}, \bibinfo{author}{G.~E. Hinton},
\newblock \bibinfo{title}{Rectified linear units improve restricted boltzmann
  machines},
\newblock in: \bibinfo{booktitle}{Proceedings of the 27th International
  Conference on Machine Learning (ICML)}, \bibinfo{year}{2010}, pp.
  \bibinfo{pages}{807--814}.
\bibitem[{Williams and Zipser(1989)}]{williams1989learning}
\bibinfo{author}{R.~J. Williams}, \bibinfo{author}{D.~Zipser},
\newblock \bibinfo{title}{A learning algorithm for continually running fully
  recurrent neural networks},
\newblock \bibinfo{journal}{Neural computation} \bibinfo{volume}{1}
  (\bibinfo{year}{1989}) \bibinfo{pages}{270--280}.
\bibitem[{Cho et~al.(2014)Cho, van Merri{\"e}nboer, Bahdanau, and
  Bengio}]{cho2014properties}
\bibinfo{author}{K.~Cho}, \bibinfo{author}{B.~van Merri{\"e}nboer},
  \bibinfo{author}{D.~Bahdanau}, \bibinfo{author}{Y.~Bengio},
\newblock \bibinfo{title}{On the properties of neural machine translation:
  Encoder-decoder approaches},
\newblock \bibinfo{journal}{arXiv preprint arXiv:1409.1259}
  (\bibinfo{year}{2014}).
\bibitem[{Chung et~al.(2014)Chung, Gulcehre, Cho, and
  Bengio}]{chung2014empirical}
\bibinfo{author}{J.~Chung}, \bibinfo{author}{C.~Gulcehre},
  \bibinfo{author}{K.~Cho}, \bibinfo{author}{Y.~Bengio},
\newblock \bibinfo{title}{Empirical evaluation of gated recurrent neural
  networks on sequence modeling},
\newblock \bibinfo{journal}{arXiv preprint arXiv:1412.3555}
  (\bibinfo{year}{2014}).
\bibitem[{Srivastava and Salakhutdinov(2012)}]{srivastava2012multimodal}
\bibinfo{author}{N.~Srivastava}, \bibinfo{author}{R.~R. Salakhutdinov},
\newblock \bibinfo{title}{Multimodal learning with deep boltzmann machines},
\newblock in: \bibinfo{booktitle}{Advances in neural information processing
  systems}, \bibinfo{year}{2012}, pp. \bibinfo{pages}{2222--2230}.
\bibitem[{Bastien et~al.(2012)Bastien, Lamblin, Pascanu, Bergstra, Goodfellow,
  Bergeron, Bouchard, Warde-Farley, and Bengio}]{bastien2012theano}
\bibinfo{author}{F.~Bastien}, \bibinfo{author}{P.~Lamblin},
  \bibinfo{author}{R.~Pascanu}, \bibinfo{author}{J.~Bergstra},
  \bibinfo{author}{I.~Goodfellow}, \bibinfo{author}{A.~Bergeron},
  \bibinfo{author}{N.~Bouchard}, \bibinfo{author}{D.~Warde-Farley},
  \bibinfo{author}{Y.~Bengio},
\newblock \bibinfo{title}{Theano: new features and speed improvements},
\newblock \bibinfo{journal}{arXiv preprint arXiv:1211.5590}
  (\bibinfo{year}{2012}).
\bibitem[{Chollet(2015)}]{chollet2015keras}
\bibinfo{author}{F.~Chollet},
\newblock \bibinfo{title}{Keras: Theano-based deep learning library},
\newblock \bibinfo{journal}{Code: https://github. com/fchollet. Documentation:
  http://keras. io}  (\bibinfo{year}{2015}).

\end{thebibliography}
	
    \clearpage
    \appendix
	\section{Appendix}
	\label{sec:appendix}
	\subsection{Feature Set C}
\label{sec:feature_set_C}
Table~\ref{tab:features_100} lists features in Feature set C.

{\small
	\begin{longtable}{ll}
	\caption{List of 135 features in feature set C.}\\
    \toprule
		\textbf{Feature Name}                         & \textbf{Table Name}    \\
    \midrule
    \endhead
		Albumin 5\%                          & inputevents   \\
		Fresh Frozen Plasma                  & inputevents   \\
		Lorazepam (Ativan)                   & inputevents   \\
		Calcium Gluconate                    & inputevents   \\
		Midazolam (Versed)                   & inputevents   \\
		Phenylephrine                        & inputevents   \\
		Furosemide (Lasix)                   & inputevents   \\
		Hydralazine                          & inputevents   \\
		Norepinephrine                       & inputevents   \\
		Magnesium Sulfate                    & inputevents   \\
		Nitroglycerin                        & inputevents   \\
		Insulin - Regular                    & inputevents   \\
		Morphine Sulfate                     & inputevents   \\
		Potassium Chloride                   & inputevents   \\
		Packed Red Blood Cells               & inputevents   \\
		Gastric Meds                         & inputevents   \\
		D5 1/2NS                             & inputevents   \\
		LR                                   & inputevents   \\
		Solution                             & inputevents   \\
		Sterile Water                        & inputevents   \\
		Piggyback                            & inputevents   \\
		OR Crystalloid Intake                & inputevents   \\
		PO Intake                            & inputevents   \\
		GT Flush                             & inputevents   \\
		KCL (Bolus)                          & inputevents   \\
		Magnesium Sulfate (Bolus)            & inputevents   \\
		epinephrine                          & inputevents   \\
		vasopressin                          & inputevents   \\
		dopamine                             & inputevents   \\
		midazolam                            & inputevents   \\
		fentanyl                             & inputevents   \\
		propofol                             & inputevents   \\
    \cmidrule{1-2}
		Gastric Tube	                     & outputevents  \\
		Stool Out Stool                      & outputevents  \\
		Urine Out Incontinent                & outputevents  \\
		Ultrafiltrate 			             & outputevents  \\
		Fecal Bag                            & outputevents  \\
		Chest Tube \#1                       & outputevents  \\
		Chest Tube \#2                       & outputevents  \\
		Jackson Pratt \#1                    & outputevents  \\
		OR EBL                               & outputevents  \\
		Pre-Admission                        & outputevents  \\
		TF Residual                          & outputevents  \\
		urinary\_output\_sum                 & outputevents  \\
	\cmidrule{1-2}
		HEMATOCRIT                           & labevents     \\
		PLATELET COUNT                       & labevents     \\
		HEMOGLOBIN                           & labevents     \\
		MCHC                                 & labevents     \\
		MCH                                  & labevents     \\
		MCV                                  & labevents     \\
		RED BLOOD CELLS                      & labevents     \\
		RDW                                  & labevents     \\
		CHLORIDE                             & labevents     \\
		ANION GAP                            & labevents     \\
		CREATININE                           & labevents     \\
		GLUCOSE                              & labevents     \\
		MAGNESIUM, TOTAL                     & labevents     \\
		CALCIUM                              & labevents     \\
		PHOSPHATE                            & labevents     \\
		INR(PT)                              & labevents     \\
		PT                                   & labevents     \\
		PTT                                  & labevents     \\
		LYMPHOCYTES                          & labevents     \\
		MONOCYTES                            & labevents     \\
		NEUTROPHILS                          & labevents     \\
		BASOPHILS                            & labevents     \\
		EOSINOPHILS                          & labevents     \\
		PH                                   & labevents     \\
		BASE EXCESS                          & labevents     \\
		CALCULATED TOTAL CO2                 & labevents     \\
		PCO2                                 & labevents     \\
		SPECIFIC GRAVITY                     & labevents     \\
		LACTATE                              & labevents     \\
		ALANINE AMINOTRANSFERASE (ALT)       & labevents     \\
		ASPARATE AMINOTRANSFERASE (AST)      & labevents     \\
		ALKALINE PHOSPHATASE                 & labevents     \\
		ALBUMIN                              & labevents     \\
		pao2                                 & labevents     \\
		serum\_urea\_nitrogen\_level         & labevents     \\
		white\_blood\_cells\_count\_mean     & labevents     \\
		serum\_bicarbonate\_level\_mean      & labevents     \\
		sodium\_level\_mean                  & labevents     \\
		potassium\_level\_mean               & labevents     \\
		bilirubin\_level                     & labevents     \\
		hgb                                  & labevents     \\
		chloride                             & labevents     \\
		peep                                 & labevents     \\
    \cmidrule{1-2}
		Aspirin                              & prescriptions \\
		Bisacodyl                            & prescriptions \\
		Docusate Sodium                      & prescriptions \\
		Humulin-R Insulin                    & prescriptions \\
		Metoprolol Tartrate                  & prescriptions \\
		Pantoprazole                         & prescriptions \\
    \cmidrule{1-2}
		ArterialBloodPressurediastolic       & chartevents   \\
		ArterialBloodPressuremean            & chartevents   \\
		RespiratoryRate                      & chartevents   \\
		AlarmsOn                             & chartevents   \\
		MinuteVolumeAlarm-Low                & chartevents   \\
		Peakinsp.Pressure                    & chartevents   \\
		PEEPset                              & chartevents   \\
		MinuteVolume                         & chartevents   \\
		TidalVolume(observed)                & chartevents   \\
		MinuteVolumeAlarm-High               & chartevents   \\
		MeanAirwayPressure                   & chartevents   \\
		CentralVenousPressure                & chartevents   \\
		RespiratoryRate(Set)                 & chartevents   \\
		PulmonaryArteryPressuremean          & chartevents   \\
		O2Flow                               & chartevents   \\
		Glucosefingerstick                   & chartevents   \\
		HeartRateAlarm-Low                   & chartevents   \\
		PulmonaryArteryPressuresystolic      & chartevents   \\
		TidalVolume(set)                     & chartevents   \\
		PulmonaryArteryPressurediastolic     & chartevents   \\
		SpO2DesatLimit                       & chartevents   \\
		RespAlarm-High                       & chartevents   \\
		SkinCare                             & chartevents   \\
		gcsverbal                            & chartevents   \\
		gcsmotor                             & chartevents   \\
		gcseyes                              & chartevents   \\
		systolic\_blood\_pressure\_abp\_mean & chartevents   \\
		heart\_rate                          & chartevents   \\
		body\_temperature                    & chartevents   \\
		fio2                                 & chartevents   \\
		ie\_ratio\_mean                      & chartevents   \\
		diastolic\_blood\_pressure\_mean     & chartevents   \\
		arterial\_pressure\_mean             & chartevents   \\
		spo2\_peripheral                     & chartevents   \\
		glucose                              & chartevents   \\
		weight                               & chartevents   \\
		height                               & chartevents   \\
    \bottomrule
    \label{tab:features_100}
	\end{longtable}
}

\subsection{Mortality Prediction Task Labels}
The labels of in-hospital mortality are derived from table \textit{ADMISSION}, in which the column \textit{DEATHTIME} records either a valid death time of an admission if the patient dies in hospital or a null value if the patient dies after discharge. Therefore, we assign the in-hospital mortality label of an admission to 1 if its \textit{DEATHTIME} is not null, else we assign the label to 0.

The labels of short-term mortality are generated with values in column \textit{INTIME} from table \textit{ICUSTAY}, which are in-time records of icu stays, and values in column \textit{DOD} from table \textit{PATIENTS}, which are records of death time of patients. We calculate the length of time interval between \textit{INTIME} and \textit{DOD} of an admission and assign its labels by comparing it with pre-defined lengths.

The labels of long-term mortality are generated with values in column \textit{DISCHTIME} from table \textit{ADMISSION}, which are in-time records of icu stays, and values in column \textit{DOD} from table \textit{PATIENTS}, which are records of death time of patients. We calculate the length of time interval between \textit{INTIME} and \textit{DOD} of an admission and assign its labels by comparing it with pre-defined lengths.

\clearpage
\subsection{MMDL model}
\begin{figure}[!h]
	\centering
	\includegraphics[width=0.9\columnwidth]{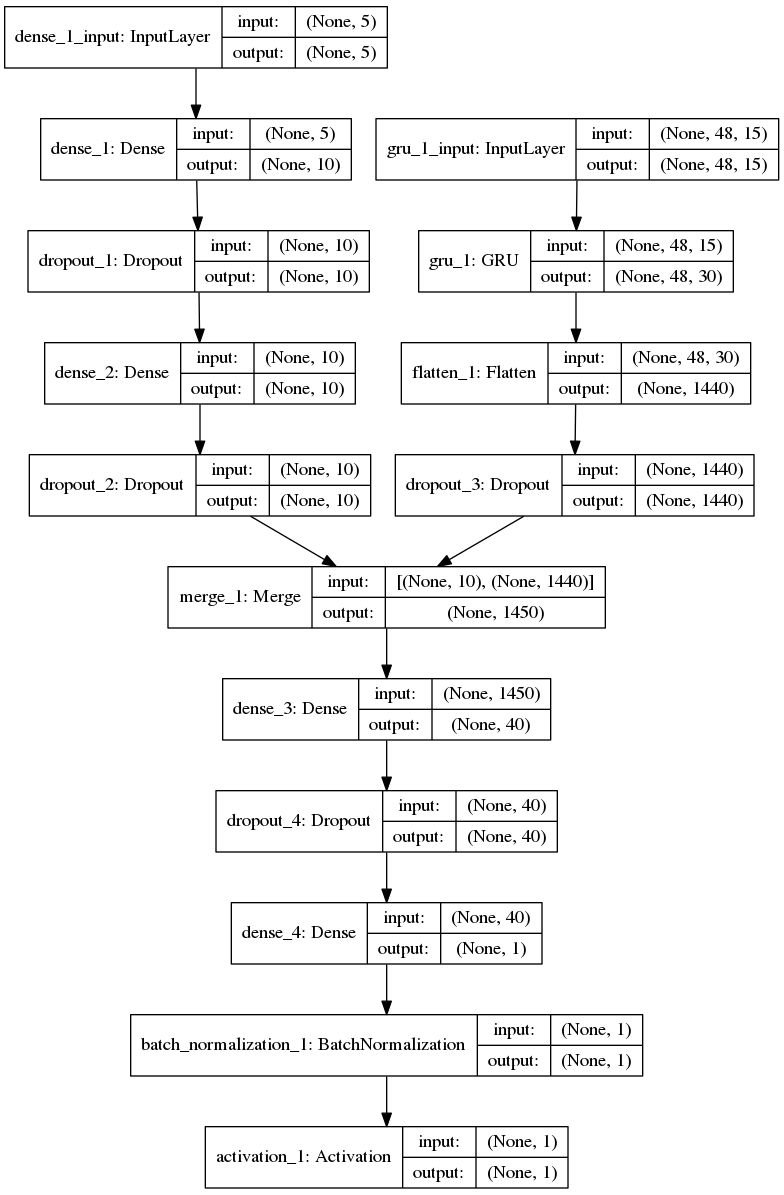}
	\caption{Structure of the MMDL model with Feature Set B as input.}
	\label{fig:ffn_lstm_struct}
\end{figure}

\end{document}